\RequirePackage{fix-cm}

\documentclass[twocolumn]{svjour3}          
\pdfoutput=1
\smartqed  

\usepackage[pdftex]{graphicx}
\graphicspath{{./Figures/}}
\DeclareGraphicsExtensions{.pdf,.jpg,.png}

\usepackage[misc,geometry]{ifsym}  
\usepackage{color}
\usepackage{footnote}
\usepackage{slashbox}
\usepackage{pict2e}
\usepackage[hidelinks]{hyperref}
\usepackage{dblfloatfix}
\usepackage{cite}
\usepackage{microtype}
\usepackage[sort&compress, numbers]{natbib}

 \journalname{Machine Vision and Applications}

\begin{document}

\title{TabletGaze: Unconstrained Appearance-based Gaze Estimation in Mobile Tablets}

\author{Qiong~Huang \and
    Ashok~Veeraraghavan \and Ashutosh~Sabharwal
}

\authorrunning{Q. Huang et al.} 

\institute{Q.~Huang (\Letter) \at
	 ECE Department, Rice University, Houston, TX, USA \\
	\email{qh3@rice.edu}
	\and
	A.~Veeraraghavan \at
	ECE Department, Rice University, Houston, TX, USA\\
	\email{vashok@rice.edu}
	\and
	A.~Sabharwal \at
	ECE Department, Rice University, Houston, TX, USA\\
	\email{ashu@rice.edu} 
}

\date{Received: date / Accepted: date}

\maketitle

\begin{abstract}
We study gaze estimation on tablets; our key design goal is uncalibrated gaze estimation using the front-facing camera during natural use of tablets, where the posture and method of holding the tablet is not constrained. We collected the first large unconstrained gaze dataset of tablet users, labeled Rice TabletGaze dataset. The dataset consists of 51 subjects, each with 4 different postures and 35 gaze locations. Subjects vary in race, gender and in their need for prescription glasses, all of which might impact gaze estimation accuracy. Driven by our observations on the collected data, we present a TabletGaze algorithm for automatic gaze estimation using multi-level HoG feature and Random Forests regressor. The TabletGaze algorithm achieves a mean error of 3.17 cm. We perform extensive evaluation on the impact of various factors such as dataset size, race, wearing glasses and user posture on the gaze estimation accuracy and make important observations about the impact of these factors. 
\keywords{Eye \and Gaze Estimation/Tracking \and Dataset \and Mobile Device \and Applications}
	
\end{abstract}

\section{Introduction} \label{sec:introduction}

Tablets are now a commonplace connected mobile computing device, and are in use worldwide for diverse applications. Current user-tablet interactions are mainly enabled by touch and sound. However, gaze is an emerging proxy of the user's attention and intention \cite{gazeAttentionCue}. Gaze information has the potential to enable a wide array of useful applications on tablets, including: i) hands-free human device interaction, such as using gaze to control the device when certain regions of the screen are hard to reach \cite{gazeForHandHeldDevice}; ii) behavior studies, such as using gaze path information for understanding and tracking reading behavior \cite{readingTrackingGaze}; and iii) user authentication when gaze-based feature is used as a biometric \cite{gazeBiometric}. In the future, many other applications could be enabled by gaze tracking on tablets.

In this paper, we study gaze estimation on the current generation of tablets, without requiring any additional hardware. Nearly all modern tablets include front-facing cameras. Our approach will be to leverage images from the front-facing cameras for gaze estimation and tracking (gaze estimation at frame rate), thereby making the resulting system suitable for today's tablets.

We adopt an appearance-based gaze estimation approach, since it does not need a calibration stage often required by many existing approaches \cite{irisTracking,brolly2004implicit,ohno2002freegaze}. Appearance-based methods find a regression mapping from the appearance of eye-region images to the gaze direction, which is then be applied to new unseen eye images. In this way, a regression model could be trained off-line, and then loaded on any tablet, estimating gaze using recorded images for any user.

A key challenge in tablet gaze tracking is the ability to robustly handle unconstrained use of tablets. During user-tablet interaction, there is often head motion, hand movement and change of body posture. As a result, shifts in the viewing angle, changes of distance between the user and the screen, and variations in illumination are possible. Moreover, any useful method should also be capable of tolerating variations in features of subject population, such as eye shape, skin and iris color, wearing glasses or not and so on. To handle the challenges, the mobile gaze tracking algorithm should be free of three constraints: i) no constraint on how people use the tablet; ii) no constraint on what kind of body posture people have when using the tablet; and iii) no constraint on the user of the tablet. 

\textit{While unconstrained gaze estimation is practically very useful, there exist no standard datasets to evaluate the reliability and accuracy of gaze estimation algorithms.}

We study the unconstrained mobile gaze estimation problem in three steps. First, we collected an unconstrained mobile gaze dataset of tablet users from 51 subjects. We name the dataset \emph{Rice TabletGaze dataset}. To the best of our knowledge, this dataset is the first of its kind and is released online for research community (\emph{\url{http://sh.rice.edu/tablet_gaze.html}}). While the dataset is collected with one tablet, gaze estimation models trained from this dataset are applicable to other handheld devices, by learned mapping between device specifications such as camera location on the tablet. The dataset consists of video sequences that were recorded by the tablet front-facing camera while subjects were looking at a dot appearing randomly on tablet screen at one of the 35 predefined locations. Subjects in the dataset are of diverse ethnic backgrounds, and 26 of them wear prescription glasses. During the data collection process, subject motion was not restricted, and each subject performed four body postures: standing, sitting, slouching, and lying. Due to our protocol design, natural and realistic subject appearance variations are captured in the dataset. We obtain a subset of our full dataset, consisting of around 100,000 images from 41 subjects. The subset is labeled with ground truth 2D gaze locations (x and y coordinates on the tablet screen), and used extensively in this paper.

\begin{figure}[!t]
	\centering
	\includegraphics[width=0.9\columnwidth]{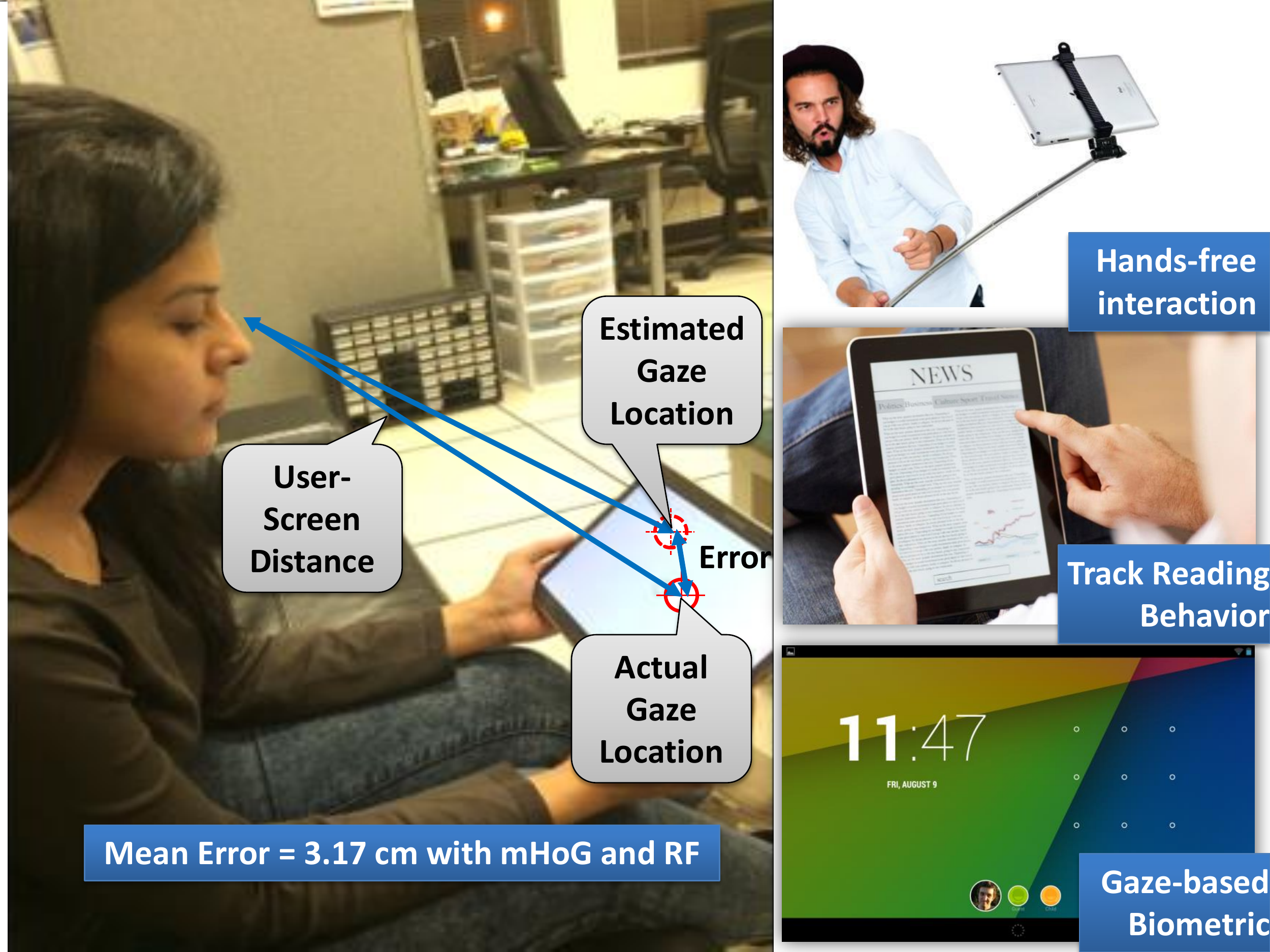}
	\caption{In this work, we provide the first dataset and an automatic algorithms for unconstrained gaze estimation on tablets. The mean error we obtained is indicated in the figure. A variety of useful applications can be enabled through gaze estimation on mobile device.}
	\label{Overview}
\end{figure}

We also present the \emph{TabletGaze algorithm} to estimate a user's gaze given an image recorded by the tablet front camera. The appearance-based TabletGaze algorithm is composed of standard computer vision building blocks. In the algorithm, the eyes in the image are first detected by a cascade eye detector \cite{eyeDetectorYu}, and then a tight region around the eyes is cropped. A multi-level HoG (mHoG) \cite{mHoG} feature is then extracted from the cropped eye images, and Linear Discriminant Analysis (LDA) is applied subsequently to reduce the feature dimensionality. The final feature is fed into a Random Forests (RF) \cite{randomForest} regressor, which outputs the location on the tablet screen at which the person in the image is gazing. The optimal combination of eye-region feature (mHoG) and regression model (RF) is found through performance comparison of 5 different features and 4 regressors on the Rice TabletGaze dataset. Then we evaluate the algorithm's performance through extensive experiments on the Rice TabletGaze dataset. The algorithm is evaluated on both person-independent and person-dependent training scenarios. We also extensively evaluate and analyze the impact of factors that could affect gaze estimation accuracy, including dataset size, race, prescription glasses and user posture. Lastly, we applied the algorithm to videos in the dataset to show continuous tracking results and demonstrated that the error variance can be reduced by using a bilateral filter. An overview of the gaze estimation system setup, the average result, and applications of gaze estimation are shown in Fig. \ref{Overview}.

In summary, this paper makes three key \textbf{contributions}:

i) Rice TabletGaze DataSet: a large gaze dataset was collected in an unconstrained mobile environment, capturing natural and realistic subject appearance variations. This dataset is publicly available at \emph{\url{http://sh.rice.edu/tablet_gaze.html}} for research purposes.

ii) TabletGaze Algorithm: An automatic gaze estimation algorithm is presented, and achieves a mean error (ME) of 3.17 cm on the tablet screen, which is significant improvement over prior art.

iii) Analysis: The study of the impact of training data size shows that the estimation accuracy can be further increased by collecting more data. We also show that for a large training dataset, dividing the dataset based on racial characteristics and body postures could improve the overall accuracy. However, partitioning the dataset based on whether or not the subject is wearing eyeglasses does not change the algorithm's performance.

\section{Related Works} \label{Related Works}
We focus on estimating the 2D location on the tablet screen where the user's eyes are focused instead of 3D gaze direction in space. A detailed summary of gaze direction estimation can be found in the following review paper \cite{gazeSurveyEyeBeholder}.

\subsection{Point of Gaze Estimation for Stationary Displays} \label{Point of Gaze Estimation for Stationary Displays}
Gaze estimation methods are typically categorized into two main groups: geometry-based, and appearance-based \cite{gazeSurveyEyeBeholder}. Geometry-based methods rely on the tracking of certain eye features, such as the iris \cite{irisTracking}, pupil center \cite{pupilTrack2,pupilTrack}, or Purkinje images \cite{glintsTrack}. To robustly track the features, those methods require extra infra-red illumination source(s), multiple cameras with calibration, and sometimes session-dependent personal calibration. 

\subsubsection{Geometry-based Methods} \label{Geometry-based Methods}
Geometry-based methods utilize explicit 3D eye ball models along with the tracked eye features to estimate the 3D gaze direction. The point of gaze is then found through the intersection of gaze direction and the screen. Based on the pupil center and Purkinje image from one camera and an infrared LED array, a double ellipse fitting mechanism was proposed in \cite{ohno2002freegaze} to predict the gaze. 
However, the system required a fixed distance between the display and the user, and head motion was limited to a 4-cm-square area. Meanwhile, an approach free of user calibration was presented in \cite{calibrationFree}. Two cameras and two point light sources that were calibrated and not co-linear were used to find the 3D locations of the cornea and pupil centers. The gaze direction was computed by connecting the cornea center and pupil center. Another approach, proposed in \cite{oneEye}, used a single image of one eye to estimate gaze direction. The iris contour in the image was modeled using an ellipse. The ellipse was then back projected into an iris circle, whose normal was regarded as the gaze direction.

\subsubsection{Appearance-based Methods} \label{Appearance-based Methods}
Appearance-based methods \cite{ALRegression, neuralNets} treat the eye region image or features extracted from the eye region image as a high dimensional vector, and learn a regression mapping model from such vector to the point of gaze (or gaze direction) through labeled training data. Such methods have the potential to be non-intrusive, free of calibration and can operate free of external hardware. A variety of regression models were utilized to find the mapping from the eye appearance to point of gaze (or gaze direction) in different works. In \cite{tanLinearManifold}, eye images were modeled as an appearance manifold. The gaze direction of a new sample was obtained from a linear interpolation of neighboring samples in the manifold model. This method was evaluated only on three subjects with fixed head pose. It used leave-one-image-out cross validation so a test subject's data appeared in the training phase. On the other hand, in \cite{GPRgaze}, a sparse, semi-supervised Gaussian Process Regression model was applied to deal with partially labeled gaze data, and realized real-time prediction of gaze direction. The method was evaluated using test images corresponding to unseen gaze locations in the training images. However, there was no description on whether a subject's data appear both in the training and testing process. In addition, there was also no description on whether the data was collected from subjects with a fixed head pose. 

In some works\cite{headPoseFree,learningBySynthesis}, 3D head pose information is extracted from images to compensate for head motion and improve gaze estimation accuracy. A two step scheme was introduced in \cite{headPoseFree} to estimate gaze direction under free head motion. The method first estimated an initial gaze direction from computed eye features under a fixed head pose, then corrected the gaze direction based on head pose rotation and eye appearance distortion. The method was also only evaluated for person-dependent scenario. In \cite{learningBySynthesis}, the authors collected a large gaze dataset with multiple head poses. Synthesized eye images were generated through 3D reconstruction of the eye region to provide more data for denser viewing angles. Then, a gaze estimation model was trained using random forest on the synthesized images. Finally, a person-independent evaluation was performed on the dataset. 

In the meantime, several datasets were released to the public for stationary displays. In \cite{gazeLocking}, Smith et al. introduced a gaze dataset composed of 5,880 images from 56 subjects. The images were recorded from a fixed distance to the subjects in a controlled environment, while they looked at each one of 21 pre-defined gaze locations. The gaze directions were coarsely arranged in seven horizontal by five vertical angles. Though five horizontal head poses were captured, the vertical head pose was fixed. Sugano et al. \cite{learningBySynthesis} collected a large dataset with 64,000 images from 50 subjects. The images have a much denser sampling of gaze angles, with 16 horizontal and 8 vertical gaze directions, and eight head poses. The images were also collected from a fixed distance to the subjects in a controlled environment. A benchmark dataset was proposed in \cite{eyediap} for evaluation of the performance of different gaze tracking/estimation algorithms. The dataset contains videos recorded by both color and depth cameras, and features the variation in head pose, type of gazing target, and ambient condition. However, the dataset included data from only 16 subjects, and only 3 subjects' data was recorded in two different ambient conditions. While all of the above mentioned datasets captured extensive amounts of head poses and appearances, the experiments were conducted in a tightly controlled manner and do not vary in body posture, which is different from our dataset that is more specifically targeting the mobile usage.

\subsection{Point of Gaze Estimation for Mobile Displays} \label{Point of Gaze Estimation for Mobile Displays}

Only a few works discussed gaze estimation methods for mobile devices, and most of those works were exploratory, directly applying previously presented methods to mobile devices. In \cite{gazeGesture}, the authors proposed using gaze gestures to control mobile phones, in comparison of gaze dwell duration, and showed the potential to improve gaze tracking accuracy by using gaze gestures. A commercial gaze tracker was utilized to locate the user's gaze location on the phone screen. The change of gaze locations was then converted to gaze gestures. This paper studied only the usability of gaze gestures to control mobile devices based on gaze tracking results, not gaze tracking itself.  
Nagamatsu et al. \cite{stereoMobiGaze} adopted the gaze tracking method proposed in \cite{corneaModelGaze}, utilizing two cameras and two light sources to find the 3D gaze direction on a mobile phone. A one point personal calibration was used to find the offset between the optical and the visual axis. The system was claimed to work under free hand movement, but there was no quantitative evaluation presented. 
Kunze et al. \cite{trackReading} implemented an application on mobile tablets and phones to accumulate statistics about user's reading behaviors. They compared the performance of one appearance-based and one geometry-based gaze tracking method, and reached the conclusion that both methods are highly dependent on not only the calibration phase but also the position in which the device was held. However, there was also no quantitative evaluation regarding the accuracy of the different methods. 
 In \cite{eyetab}, an on-device gaze tracking prototype was implemented using a geometry-based gaze estimation method on an unmodified tablet. The algorithm fitted an ellipse to eye limbus within the region-of-interest (ROI) detected by eye detectors, and found the optical axis through the ellipse normal vector. No user calibration was performed to correct the error between the optical and the visual axis. The optical axis was directly treated as the gaze direction. An accuracy of 6.88$^{\circ}$ was claimed in the work. However, the method was evaluated only on 8 subjects, and subject-tablet distance was fixed in the experiments. Furthermore, the gaze locations included only 9 dots on the screen, covering part of the available tablet surface. 
Recently Zhang et. al \cite{wildGazeEstimation} presented a gaze dataset collected under free laptop use with 15 participants. The dataset contains 213,659 images and has 20 gaze locations. An algorithm was also presented in the work utilizing multimodal convolutional neural networks (CNN) to predict gaze direction from head pose and eye appearance. Though laptops are technically a mobile device, they have much less mobility compared to \textit{handheld} devices like tablets and phones. In addition, the statistics from this work showed that the majority of the data was collected during work time, when people would more likely put their laptops on the desk. A major impact from this difference is that the users face is fully visible , while it is certainly not the case for tablets, as is shown in our work. Furthermore, the algorithm presented requires camera calibration and a pre-built facial shape model.

Our work is the first to study unconstrained handheld mobile device gaze estimation. Our gaze dataset was collected with free subject motion and different body postures, greatly capturing the appearance variations in unconstrained environments. In addition, our algorithm is fully automatic and is developed based on the observations made on the Rice TabletGaze dataset. Our study on the impact of practical factors on the algorithm performance such as prescription glasses and body posture, as well as our evaluation of continuous gaze tracking, help us understand mobile gaze estimation and its practicality.

\section{Rice TabletGaze Dataset} \label{Unconstrained Mobile Gaze Dataset}
We created the first publicly available unconstrained mobile gaze dataset, Rice TabletGaze Dataset, to provide data for our study of the unconstrained mobile gaze estimation problem. We designed our data collection experiments to capture unique, unrestrained characteristics in the mobile environment. To this end, we have collected data from 51 subjects, each with four different body postures. The dataset is also released online to promote future research development of unconstrained gaze estimation methods. While all the data in this paper is recorded with one tablet, one could potentially train a gaze estimation model from this dataset, and the learned model can be used for gaze estimation on other handheld devices through approaches that use transfer learning, domain adaptation or by directly encoding the relative location and resolution of the cameras in the two devices. While, we believe this is feasible, it is outside the scope of this paper.

\subsection{Data Collection} \label{Data Collection}
\begin{figure}[!t]
	\centering
	\includegraphics[width=3in]{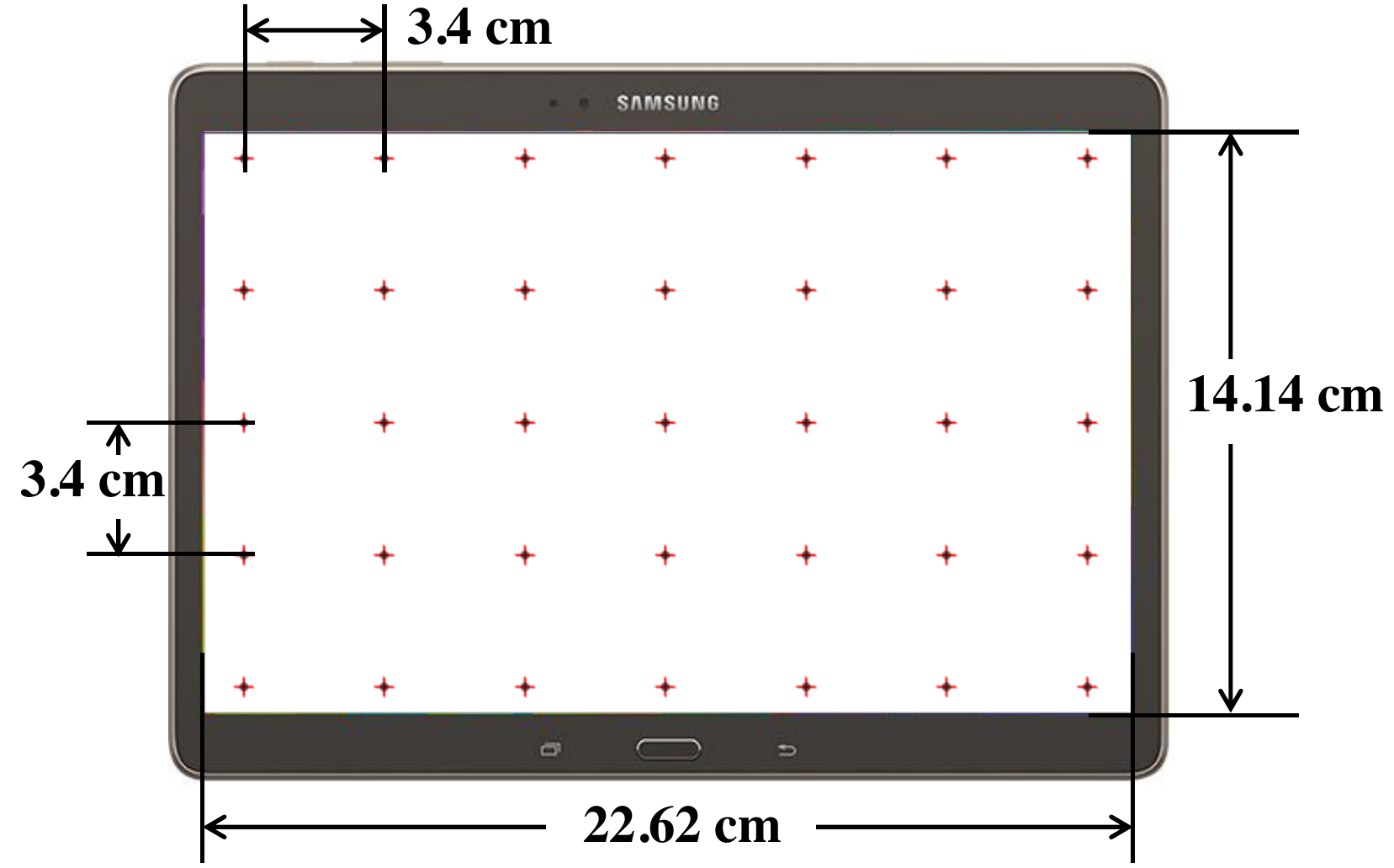}
	\caption{Gaze locations on the tablet screen. There are 35 (5 $\times$ 7) locations distributed on the tablet screen. In one data collection session, a dot appeared at one location at a time, and then moved to another location after 3 seconds. This continued until the dot had appeared at all the 35 locations once. The location of the dot was randomized among the 35 points.}
	\label{gazePoints}
\end{figure}

In this research, we used a Samsung Galaxy Tab S 10.5 tablet with a screen size of 22.62 $\times$ 14.14 cm (8.90 $\times$ 5.57 inches). A total of 35 gaze locations (points) are equally distributed on the tablet screen, arranged in 5 rows and 7 columns and spaced 3.42 cm horizontally and 3.41 cm vertically. Example images of the gaze pattern on the tablet screen is shown in Fig. \ref{gazePoints}. The raw data are videos captured by the front-camera of the tablet that was held in landscape mode by the subjects, with an image resolution of 1280 $\times$ 720 pixels.

A total of 51 subjects, 12 female and 39 male, participated in the data collection, with 26 of them wearing prescription glasses; 28 of the subjects are Caucasians, and the remaining 23 are Asians. The ages of the subjects range approximately from 20 to 40 years old. An institutional review board (IRB) approval is obtained for the research and all subjects signed a consent form to allow their data to be used in the research and released online. 

\begin{figure}[!t]
	\centering
	\includegraphics[width=\columnwidth,height=1.3in]{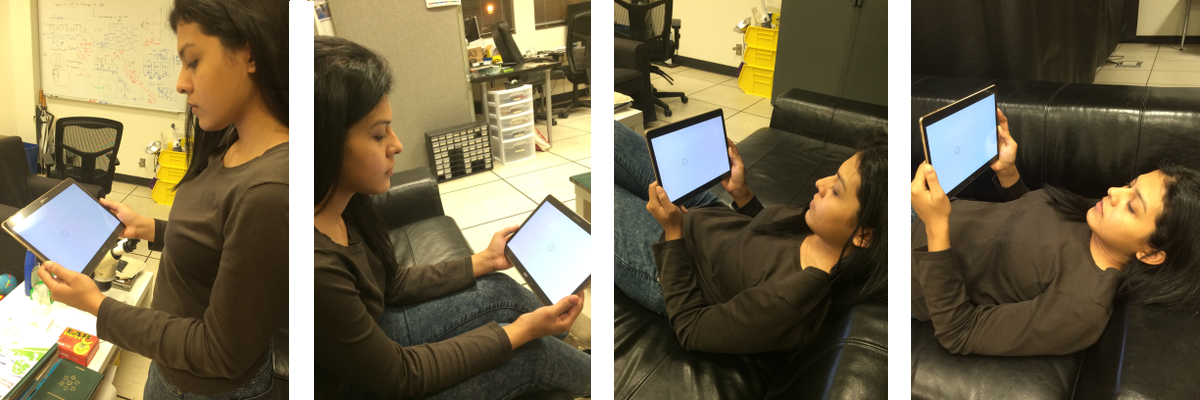}
	\caption{An example image of the data collection process. In one data collection session, a subject maintains one of four body postures while gazing at a dot on the tablet screen. At the same time, a video of the subject is recorded by the tablet front camera. From left to right, the subject is standing, sitting, slouching and lying.}
	\label{dataCollection}
\end{figure}

During each data collection session, the subject held the tablet in one of the four body postures (standing, sitting, slouching or lying) as shown in Fig. \ref{dataCollection}, and recorded one video sequence. Each subject needed to conduct four recording sessions for each of the four body postures, so a total of 16 video sequences were collected for each subject. For each recording session, there was no restriction on how the subject held the tablet or how they performed each body posture. The data collection happened in a naturally lit office environment, where only the ceiling lights directly on top of the subjects were turned off to reduce the strong background light in the recorded videos.

\begin{figure*}[!t]
	\centering
	\includegraphics[width=6in]{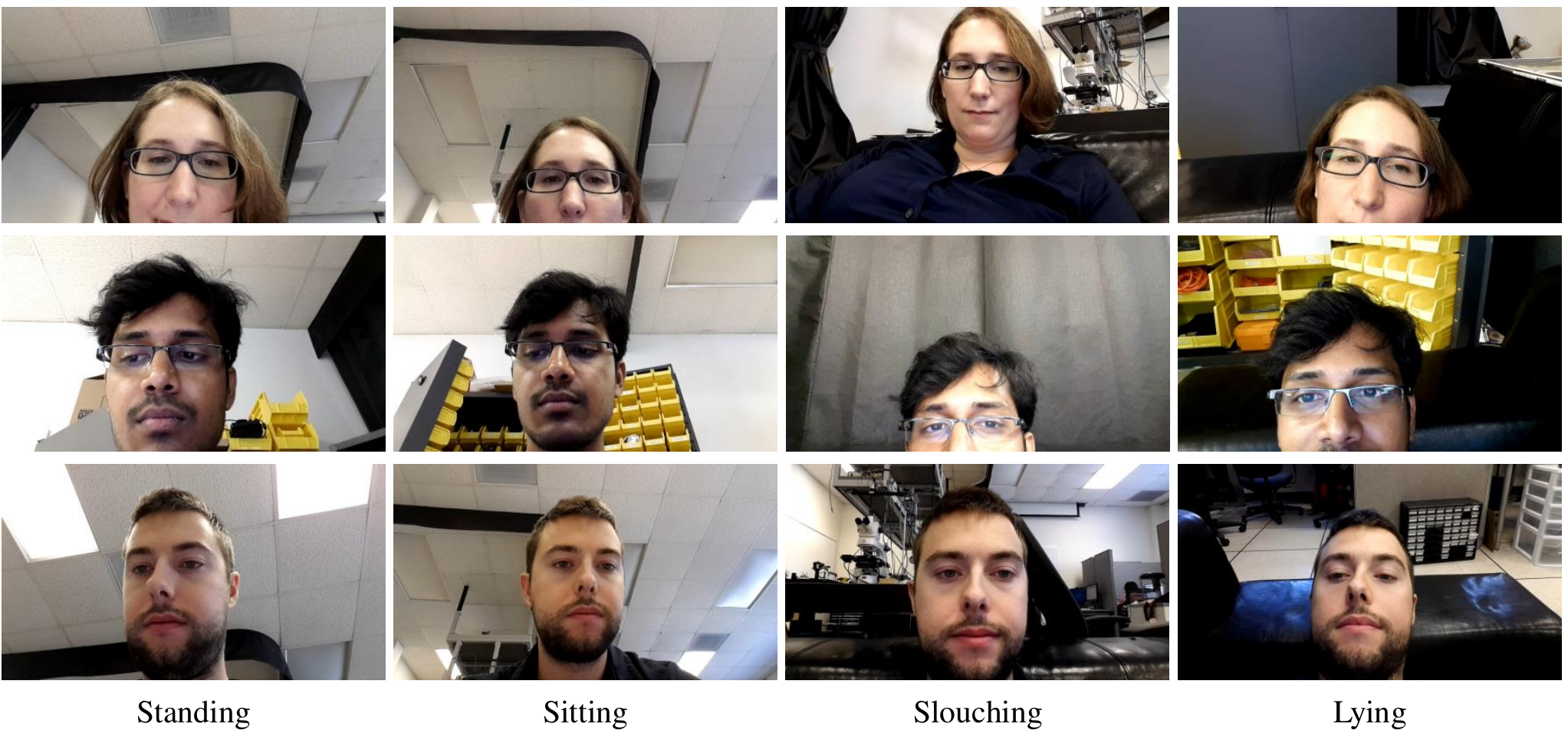}
	\caption{Sample images from the Rice TabletGaze dataset. We observe subject appearance variations across different recording sessions. Also, notice that only a fraction of the subject's face is visible in a fraction of the images in the dataset.}
	\label{imageDataset}
\end{figure*}

When a subject started one data collection session, he or she initialized a background recording application on the tablet, so the front facing camera of the tablet began recording a video of the subject with audio. Then the subject started to play and watch a video on the tablet. A beep sound notified the beginning of the video, which was also recorded in the video sequence. The recorded sound would be utilized later to locate the time instant in the recorded video when the subject started to watch the video. The video watched by the subjects consists of a dot changing its location every three seconds, and the subject was instructed to focus his/her eyes on the dot the whole time. The subject was free to blink his/her eyes, as it would be uncomfortable to restrain the eye blink in each approximately two minute long data collection session. To prevent the subject from focusing his eyes to the next gaze point ahead of time (i.e. predicting the dot location), the location of the dot was randomized among the 35 possible points. Sample images from the dataset are shown in Fig. \ref{imageDataset}.

\subsection{Observations on the Rice TabletGaze Dataset} \label{Data Observations}

In this section, we discuss our observations about facial visibility, body posture and prescription glasses, based on our TabletGaze dataset described in Section \ref{Data Collection}. 

\textbf{Observation 1: The entire face may not be visible in most of the image frames.}

\begin{figure}[!t]
	\centering
	\includegraphics[width=\columnwidth]{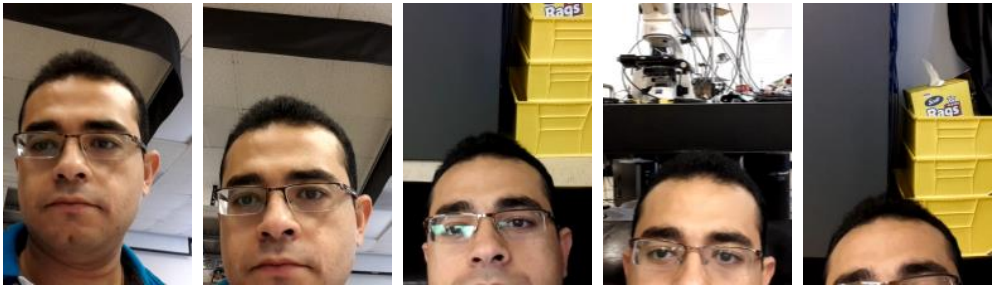}
	\caption{Example images of five different levels of facial visibility categories. From left to right, each image represents one of the five following visibility categories: i) the whole face, ii) from mouth and above, iii) from nose and above, iv) from eyes and above, and v) where even the eyes are not visible. For clarity of presentation, we have cropped the background.}
	\label{facialVisibility}
\end{figure}

Fig.~\ref{facialVisibility} shows an example of full range of facial visibility for the same subject during different data collection sessions. The images vary from full facial visibility to only the subject's forehead being visible. To quantify the extent of facial visibility, we labeled each video in the TabletGaze dataset as belonging to one of the following five categories: (i)~the whole face; (ii)~from mouth and above; (iii)~from nose and above; (iv)~from eyes and above; and (v)~even the eyes are not visible. For each video sequence, we manually reviewed 4 images (each image corresponds to 1 of the 4 corner gaze locations on the tablet screen) and determined the facial visibility extent of each image. The video sequence is labeled as the majority category of the 4 images. The statistics based on the above categorization are shown in Table~\ref{head_visible_posture_table}. 

We observe that the whole face is visible in only 30.8\% of all the videos, and the number varies from one posture to another, with sitting being the highest (47\%) and lying being the lowest (13.7\%). It is clear that in a strong majority of the videos, full facial visibility cannot be assumed. 

The extent of facial visibility directly affects the amount of information that can be extracted from the facial region for gaze estimation. For example, head pose information (pitch, yaw and roll angles) can be estimated from the face, and can be used in conjunction with eye appearance information to improve gaze estimation. The details were discussed in Section \ref{Related Works}. The bulk of previously proposed head estimation methods \cite{ASModelHeadPose, featureBasedHeadPose, mainfoldHeadPose} require the whole face to be visible, and are not effective when only part of the face is visible. 
Due to a lack of robust methods for extracting head pose estimation using partial face visibility, we largely focus on eye region appearance in this paper. However, we did perform preliminary work to incorporate implicit head pose information, such as eye locations in the image frame, as discussed in detail in Section \ref{Discussion and Conclusion}. In addition, methods that incorporate head pose information for those frames where the entire face is visible will potentially improve gaze estimation accuracy, though such an investigation is outside the scope of this paper.

\begin{table}[!t]
	\renewcommand{\arraystretch}{1.5}
	\caption{Statistics on the extent of the visible face region. Each video in the dataset is labeled as one of the five facial visibility categories. The numbers in the table are percentage of videos. Note that the whole face is only visible in 30.8\% of all the videos. Based on this data, we can infer that most of the time the whole face is not visible. }
	\label{head_visible_posture_table}
	\centering
	\resizebox{\columnwidth}{!}{
		\begin{tabular}{|c|c|c|c|c|c|}
			\hline
			\backslashbox{Posture\kern-1em}{\kern-1em Facial Visibility} & Whole & Mouth & Nose & Eyes & No Eyes\\
			\hline
			Standing & 39.2\% & 38.2\% & 18.6\% & 4.0\% & 0\% \\
			\hline
			Sitting & 47.0\% & 27.5\% & 19.1\% & 5.9\% & 0.5\% \\
			\hline
			Slouching & 23.0\% & 35.8\% & 26.0\% & 13.2\% & 2.0\% \\
			\hline
			Lying & 13.7\% & 39.7\% & 35.3\% & 7.4\% & 3.9\% \\
			\hline
			All body postures & 30.8\% & 35.2\% & 24.8\% & 7.6\% & 1.6\%\\
			\hline
		\end{tabular}	
	}
\end{table}

\textbf{Observation 2: Body posture and facial visibility extent appear to be correlated.}

Our starting hypothesis was that there might be a correlation between facial visibility extent and body posture during tablet use. Two main conclusions can be derived from Table~\ref{head_visible_posture_table}. First, when seeking a refined amount of information about facial visibility, body posture information can be useful. For example, standing/sitting postures lead to higher probabilities of the face being fully visible, compared to slouching/lying. Intuitively, the observations make sense based on practical experience. Most users tend to rest their tablets on their chest/abdomen when slouching/lying, which reduces chances of seeing the whole face. Although this is beyond the scope of this paper, facial visibility extent could thus potentially be used to roughly estimate the body posture. 

Second, if the only objective is to see the eyes, then the eyes are visible in at least 96\% of the videos for any posture. Thus, for our proposed appearance-based method discussed in Section \ref{Baseline Estimation Framework}, which relies on the visibility of the eyes only, information about body postures is not essential. However, for methods that may rely on other facial landmarks, the accuracy of gaze estimation could be dependent on the body posture.

\begin{figure}[!t]
	\centering
	\includegraphics[width=\columnwidth]{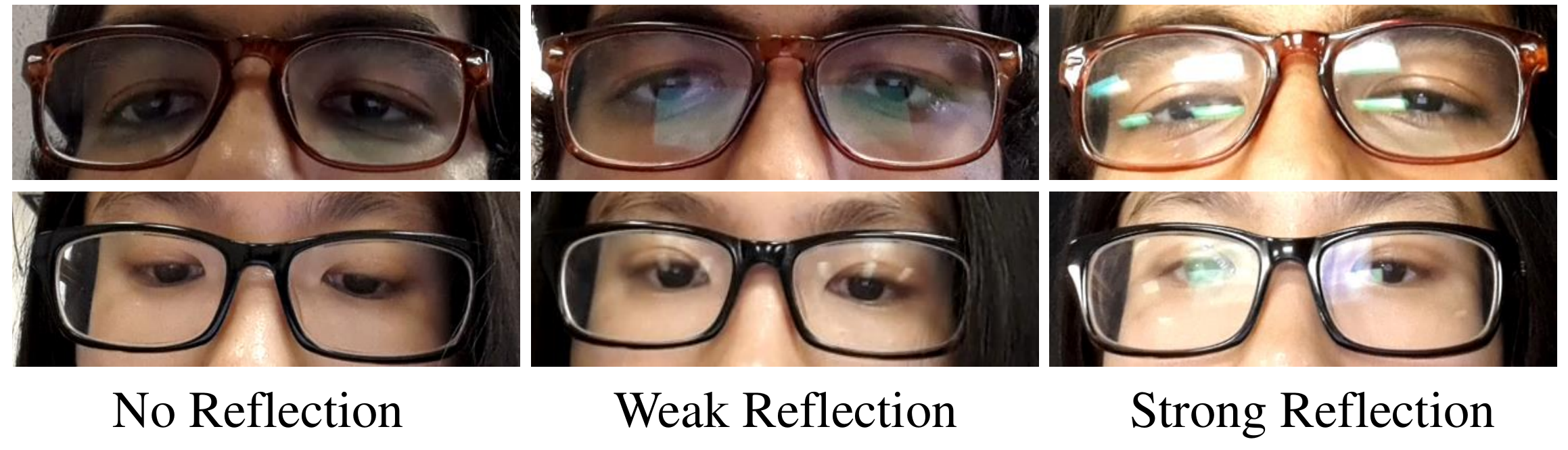}
	\caption{Example images of different glasses reflection strength. From left to right, each image represents no reflection, weak reflection and strong reflection respectively.}
	\label{reflections}
\end{figure}
\begin{table}[!t]
	\renewcommand{\arraystretch}{1.5}
	\caption{Statistics on eyeglasses reflection strength. We can infer that prescription eyeglasses cause reflection in approximately half of the videos.}
	\label{glass_reflection_table}
	\centering
	\begin{tabular}{|c|c|c|c|}
		\hline
		Reflection Strength & None & Weak & Strong \\
		\hline
		Number of videos & 49.5\% & 24.2\% & 26.3\% \\
		\hline
	\end{tabular}	
\end{table}

\textbf{Observation 3: Prescription glasses can cause reflection, and in many instances, the reflection can be significant.}

Fig.~\ref{reflections} shows examples of eyeglasses reflections from the TabletGaze dataset. Depending on the viewing angle, light source, orientation and coating, there may be no glare from the eyeglasses (left most image in Fig.~\ref{reflections}) or very strong glare (right most image in Fig.~\ref{reflections}). 

To quantify how often reflection happens and how strong the reflection is, we accumulated information on the occurrences and strength of eye glasses reflections in the eye image. We categorized the videos into three broad categories (no reflection, weak reflection and strong reflection) by the same method we used for face visibility categorization. The categorization is done for all the videos of subjects who were wearing glasses, and the statistics are listed in Table~\ref{glass_reflection_table}. We observe that there is visible glasses reflection in half the videos, and in 26.3\% of the videos, there is a strong reflection. Reflections with strong intensities could potentially impact the gaze estimation accuracy by i) possibly confusing eye detector used in our algorithms, making it return an erroneous bounding box location around the eye region, and ii) reducing the contrast in some regions of the eye, which in turn makes part of the eye, such as iris or sclera, invisible.

\begin{figure*}[!b]
	\centering
	\includegraphics[width=\textwidth]{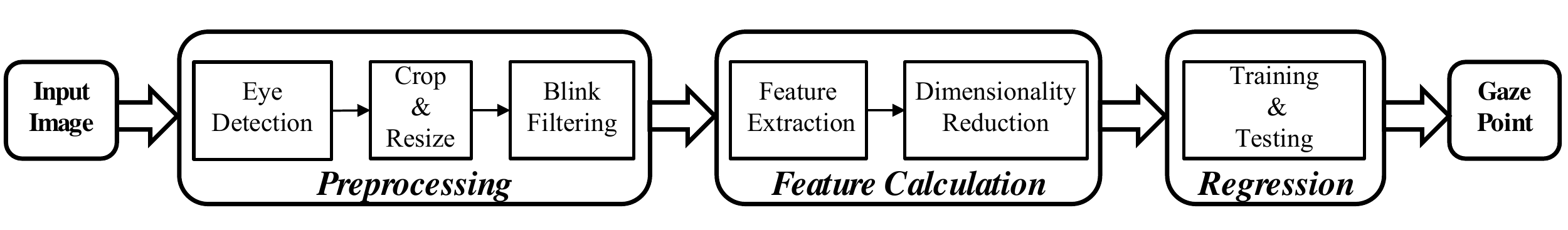}
	\caption{Automatic gaze estimation algorithm. The input to the algorithm is an image of the user recorded by the tablet front camera. The output is the location on the tablet screen at which the user is looking. The estimation of gaze from an image consists preprocessing, feature extraction and regression estimation. }
	\label{baselineFig}
\end{figure*}

\subsection{Sub-dataset Labeling} \label{Data Labeling}

The total amount of raw data collected is 51 $\times$ 16 = 816 video sequences. However, a portion of the data is not usable for three reasons: i) the transition from one gaze point to the next and loss of concentration of subjects produce image frames with inconsistent gaze labels; ii) the eye detector failure in some conditions causes missing data, and iii) involuntary eye blinks and large motion blur result in images without useful gaze information from the eyes. Because of these three reasons, we prune the raw data and obtain a sub-dataset of 41 subjects to be used in our experiments. Below, we explore the three reasons in more detail and describe how we filter out the unusable data.

We first remove images with inconsistent gaze labels. We extract only the video chunk that corresponds to 1.5 to 2.5 seconds after the time the dot appears at a new location to remove the time for subjects to re-focus. Since it is unavoidable that sometimes the subject loses concentration during a data collection session, the gaze label of parts of the corresponding video data can be mismatched. For the 35 video chunks extracted from each video sequence, we visually inspect whether there is a gaze drift for more than 5 video chunks and, if so, abandon the data from the whole video sequence. Since it is hard to determine the true gaze location just by looking at one stand-alone image, we extract one eye region image for each gaze point and enhance the contrast of the image to compensate for the low illumination scenario. By comparing the relative location of the iris and openness of the eyes among 35 gaze locations, we are able to identify each gaze drift occurrence and calculate the total number of gaze drifts.

\begin{figure}[!t]
	\centering
	\includegraphics[width=\columnwidth]{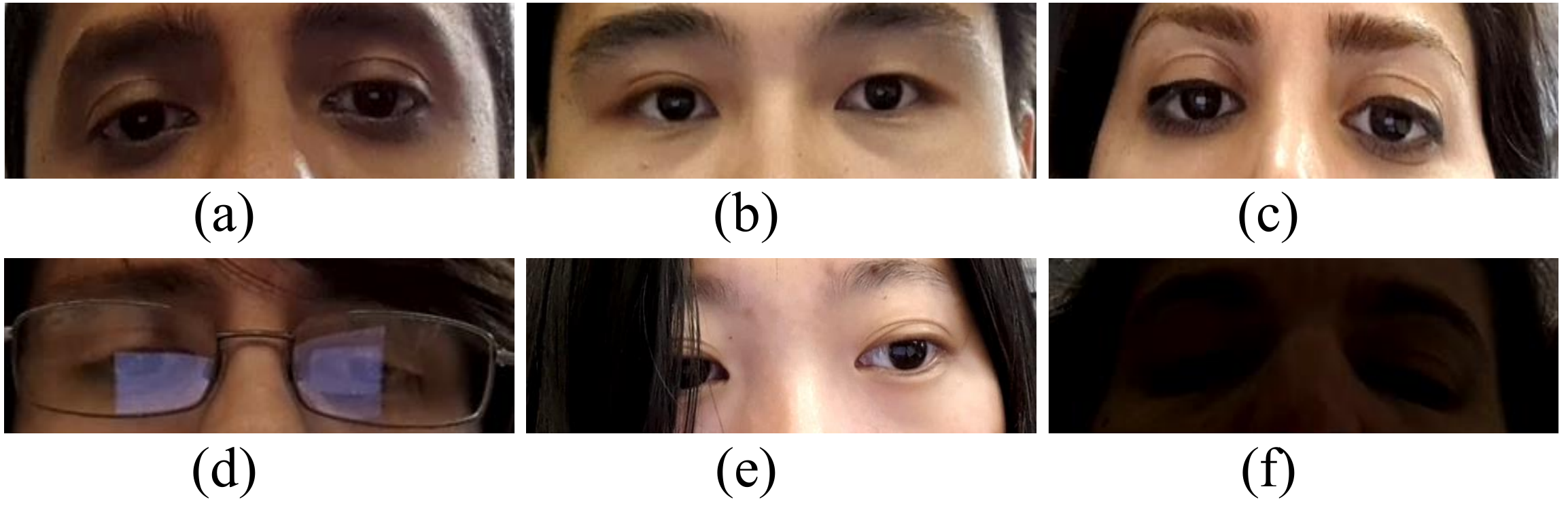}
	\caption{Eye detection fails in some scenarios. Top row of images, (a-c), show correct eye detection cases. Bottom row of images represent eye detection failure cases, including (d) strong glasses reflection, (e) hair occlusion over an eye, and (f) poor illumination.}
	\label{EyeDectionFailure}
\end{figure}

We then remove images with eye detector failures. For each video chunk of time duration 1 second, the number of frames contained is between 15 and 30 due to the variable video recording rate of the front camera. An important step for automatic estimation of gaze through images is to detect the eye region using an eye detector, which fails in conditions such as eyes are not visible in the image frame, strong reflection from prescription glasses, occlusion from hair, poor illumination, and so on. Images with eye detector failures are removed, resulting in small data size for certain subjects. Examples of eye detection success and failure cases are shown in Fig. \ref{EyeDectionFailure}.

Another source of images without useful information is the involuntary blinking and occasional large motion of the subjects during the data collection stage. Since the images of closed eyes and blurred eye regions are undesired, for image frames within each video chunk corresponding to one gaze direction, we extract 5 images with lower mean intensity value and higher mean Laplacian of Gaussian (LoG) value. We do this because images of closed eyes will have higher mean intensity value given the disappearance of the dark pupil, and a blurred eye region image will have a lower mean LoG value because motion blur weakens the edge information in the image. Even though some video chunks do not contain closed eye images, we still extract 5 image frames to guarantee a similar number of data samples for each gaze point.

This extensive data selection process removes most of the unusable images. The tiny fraction of bad images that escape this procedure is treated as noise.

\section{TabletGaze: Gaze Estimation System} \label{Baseline Estimation Framework} 

In this section, we describe the proposed TabletGaze, the gaze estimation framework that leverages well-known machine learning processing modules, as shown in Fig. \ref{baselineFig}. The estimation of gaze from an image consists of three parts: preprocessing, feature extraction and regression estimation. The preprocessing part involves image normalization (e.g. scaling) so the eyes from different images can be directly compared. For feature extraction and regression, we utilize a data-driven approach to guide the selection of features and regressors. 
We tested five features including contrast normalized intensities, Laplacian of Gaussian (LoG), Local Binary Patterns (LBP) \cite{LBP}, Histogram of Oriented Gradients (HoG) \cite{HoG}, and multilevel HoG (mHoG) \cite{mHoG}. We utilized four regressors, namely $k$-Nearest Neighbors ($k$-NN), Random Forests (RF) \cite{randomForest}, Gaussian Process Regression (GPR) \cite{GPR}, and Support Vector Regression (SVR) \cite{SVR}. 

\subsection{Preprocessing} \label{Preprocessing}
\begin{figure}[!t]
	\centering
	\includegraphics[width=\columnwidth]{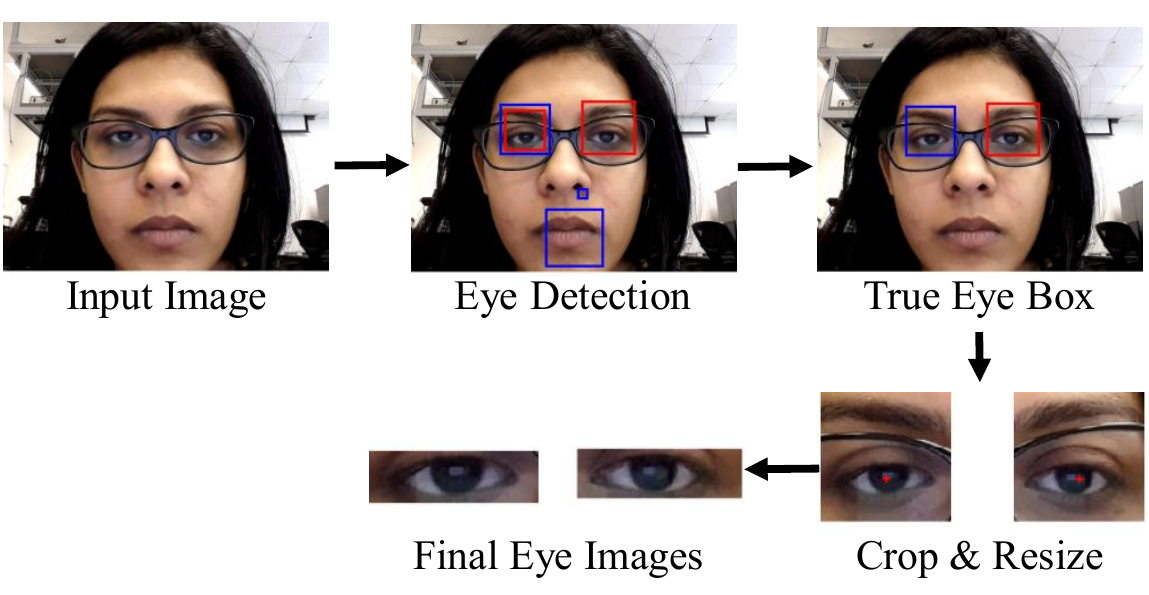}
	\caption{Example images in the preprocessing phase. Firstly, two eye detectors are applied to detect potential left and right eye regions. The blue bounding boxes denote the output of left eye detector, while the red of right eye detector. In the example image, we can observe false positive image patches around the nostril and mouth, which are removed to find the true eye region. Then only a tight region around the eyes is used to avoid the ambiguity caused by eye brows and facial expressions.}
	\label{Preprocess}
\end{figure}

\begin{figure}[!t]
	\centering
	\includegraphics[width=\columnwidth]{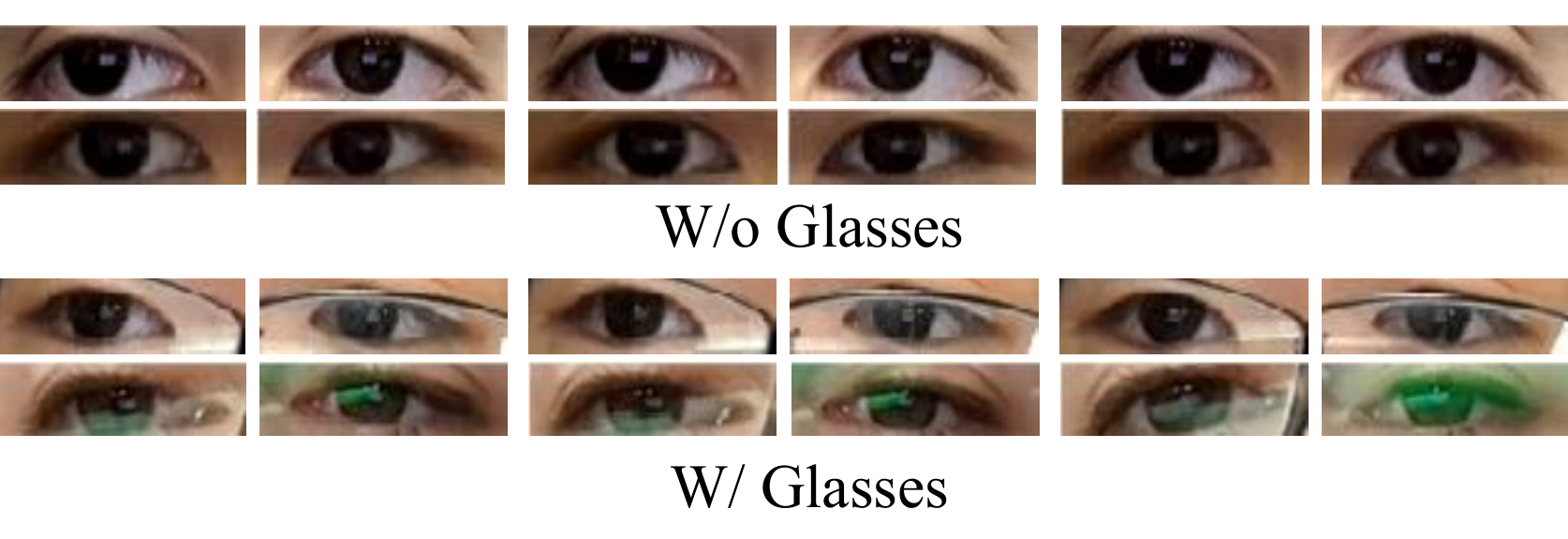}
	\caption{A few sample images of the final extracted eyes data. Each row of eye images comes from one subject. We observe that after the preprocessing step, the eyes are tightly cropped.}
	\label{processedEyes}
\end{figure}

The first step in TabletGaze is to preprocess the input images, which have a resolution of 1280 $\times$ 720 pixels. An example of the preprocessing step is displayed in Fig. \ref{Preprocess}. We first apply two Harr feature CART-tree based cascade detectors \cite{eyeDetectorYu}, one trained for left eye and one for right eye, to locate image patches that includes potential left and right eye regions. A sample output of the detectors can be found in Fig. \ref{Preprocess}. False positive bounding boxes from the detectors are rejected by 1) empirically establishing a threshold for the size of the box to remove small false-positive patches, such as the nostril detected in Fig. \ref{Preprocess}, and 2) enforcing coarsely symmetric locations of the bounding boxes returned by the left and right eye detectors (to compensate for head tilt where eyes are not totally symmetric) to remove stand alone false-positive patches, such as the mouth detected in Fig. \ref{Preprocess}. The eye region bounding box sizes vary for different images, so their sizes are scaled to 100 $\times$ 100 pixels. The detected bounding box contains a large area including the eye brows, which is not informative about gaze, so we crop a tight box around the eye to procure the final eye image. The pupil center is coarsely located at one half horizontally and two thirds vertically of the bounding box, given the aforementioned eye detector was trained with eye images of this geometry. We crop 15 pixels from the top and bottom around the pupil center to form the final eye image, which covers the eye region tightly for most subjects. The horizontal dimension is untouched since the eye width varies widely among different subjects. As a result of the aforementioned operations, the final eye image size becomes a fixed 30$\times$100 pixels for each eye across all images. A few sample images of the cropped eye regions are shown in Fig. \ref{processedEyes}. Blinks cause the gaze estimation algorithm to produce incorrect predictions and need to be removed. To detect blinks, the algorithm looks for changes in the mean pixel intensity of the eye region over time. The algorithm takes advantage of the fact that when an eye blink occurs, the continuous disappearance and reappearance of the dark pupil results in an increase, then decrease of the mean pixel intensity. The mean is taken over 20 consecutive frames, which is usually less than the time length between two consecutive blinks. By inspecting the video sequences, we found that an eye blink usually lasts around 4-6 frames. Therefore, when a blink is detected, we skip 6 frames around the peak frame.

\subsection{Feature Calculation} \label{Feature Calculation}
Following eye extraction, we next find features. Feature calculation includes two steps: feature extraction and dimensionality reduction. 

\textbf{Feature Extraction:}
The accuracy of gaze estimation greatly depends on the feature we choose. To ensure our algorithm achieved a state-of-the-art result, we chose to evaluate the performance of 5 popular features: (1) contrast normalized pixel intensities; (2) LoG; (3) LBP; (4) HoG; and (5) mHoG feature. The first proposed feature, contrast normalized pixel intensities, is the simplest feature of the five; it converts pixel values into the feature vector after normalization to account for variations in illumination. LoG convolves each eye image with a LoG filter and concatenates the returned vector to enhance eye contour and remove person-dependent eye region texture information. LBP and HoG have been proven by many works as powerful features \cite{HoGPowerful}. LBP captures image texture information, while HoG retrieves local shape and orientation information. As a variant of HoG, multilevel HoG (mHoG) is formed by concatenating HoG features at different scales. The block scales utilized in this paper are the same as presented in \cite{mHoGParams}.

\textbf{Dimensionality Reduction:}
Features obtained in the feature extraction phase suffer from being high dimensional and compromised by noise. We overcome these problems by mapping the features to a lower dimensional space. In this work, we applied Linear Discriminant Analysis (LDA) to reduce the feature dimensionality. LDA maps the data to a lower dimensional space where the inter-class scatter to intra-class scatter ratio is maximized. Finding the projection vector requires computation of the inverse of intra-class scatter. The intra-class scatter matrix suffers from a singularity problem when the number of data samples per class is smaller than the number of features. Regarding this, we applied Principal Component Analysis (PCA) to the original feature data to reduce its dimension. The dimension is reduced to no smaller than the number of observations per gaze point. Then we apply LDA to the already reduced data to obtain a final feature vector. Given input data of feature length $C$, the output data of LDA will have a length of $C-1$. In our dataset, we have gaze data corresponding to 35 gaze locations, so the final data after the LDA operation has a feature length of 34. 

\subsection{Regression} \label{Regression}
Finally, after computing the final feature vectors, the data is fed into a regression model. The gaze labels of the data include two parts: the horizontal and vertical (x and y) coordinates on the tablet screen. We trained a separate regressor for the horizontal and vertical gaze locations respectively. Then the output from the two regressors are combined as the predicted 2D gaze location on the tablet screen. In our work, we experimented with the four different models mentioned earlier. $k$-Nearest Neighbors ($k$-NN) assigns the average of the output of the $k$ nearest neighbors in the training data to a new observation; we chose $k=3$ in our experiments. 

Random Forests (RF) are a set of weak binary tree regressors. Each tree in the forest is grown by randomly boostrapping samples and each binary split of the tree is grown by randomly selecting a subset of the features. For regression RF, the output of a new input is given by the average of the output of each tree in the forest. RF has previously been used in gaze estimation papers and shows strong performance \cite{learningBySynthesis,randomForestGaze}. In our experiments, we used 100 trees.

Gaussian Process Regression (GPR) models the regression problem as a Gaussian process and estimates the output of a new observation by taking the conditional probability over the training data. The advantage of GPR is that it returns not only the estimate of the output, but also the confidence interval of the estimate. However, traditional GPR has a complexity of $O(N^{3})$ for an input data samples size $N$ \cite{GPRcomplexity}, which makes it computationally infeasible for a large dataset, such as the over 100,000 samples in our data. In our experiments, we used fully independent training conditional (FITC) approximation \cite{FITCapproximation}, a sparse GPR method which claims to achieve similar accuracy as GPR, to reduce the running time. Even with the faster FITC approximation, we could only manage to evaluate on 15 subjects with a reasonable computing time using three-fold cross validation.

Support Vector Regression (SVR) utilized the well-known "kernel trick" to project data into a higher dimensional space where a linear regression function can effectively fit the data. A nonlinear kernel can transform a non-linear regression problem in the original data space into a linear one in the new space. In our experiment, we employed the popular non-linear radial basis function (RBF) kernel. The performance of SVR depends highly on the model parameters, which are usually obtained through a coarse to fine grid-search process. Given a data sample size $N$, SVR has a training time complexity of $O(N^{3})$ \cite{SVRcomplexity}, which greatly limits its scalability to large datasets. In our experiments, we evaluated SVR on the subset of 15 subjects that was used in GPR evaluation. The evaluation was also conducted using three-fold cross validation.

\begin{figure*}[!t]
	\centering
	\includegraphics[width=6.2in]{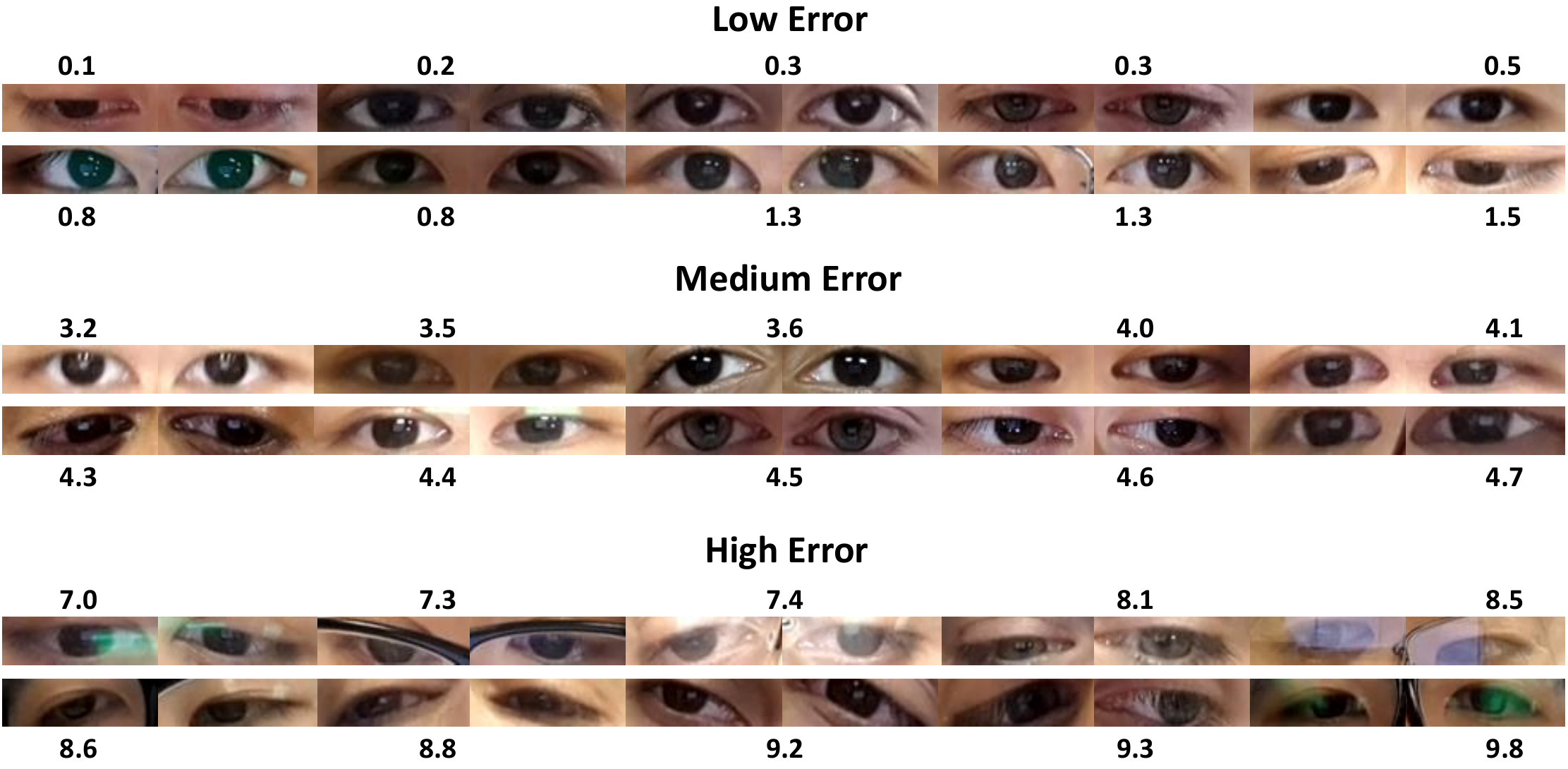}
	\caption{Example eye images with different gaze estimation errors. In the figure, we show 10 pairs of eyes for low, medium, and high estimation errors (in cm) using mHoG + RF gaze estimation algorithm. We can observe that factors such as erroneous eye region cropping, long eyelashes, strong reflections from prescription glasses, glass frames, rotated eyes, and motion blur can reduce estimation accuracy.}
	\label{badPrediction}
\end{figure*}

\section{Results and Analysis} \label{Results and Analysis}
\vspace{0.05in}
\subsection{Error Metric} \label{Error Metric}
Previous works on gaze estimation employed an angular error to evaluate the quality of gaze estimation. The angular error is computed by taking the arctangent of the ratio between the distance from the subject's eyes to the screen center and the distance of the gaze point from the screen center. However, in the mobile environment, the distance between the screen and the user is highly variable, so it is not possible to reliably calculate the angular error. For our work, since we have the ground truth gaze labels (2D location on the table screen) for all data, we define the estimation error of one data sample as the Euclidean distance between the predicted 2D gaze point location and the actual 2D gaze point location on the tablet screen. The final error is reported as the mean error (ME) over all data samples.

\subsection{Comparisons for Different Features + Regressors} \label{Comparisons for Different Features + Regressors}

In Table~\ref{resultTable}, we first summarize the performance of each feature and regressor as described in Section \ref{Feature Calculation} and \ref{Regression}. The entries in Table \ref{resultTable} are the MEs (in cm) across around 100,000 images from 41 subjects using the cross validation method described in Section \ref{Regression}. The columns of Table~\ref{resultTable} represent different features. The features are listed in order of increasing complexity, and this trend can be seen in the table - estimation accuracy generally increases as feature complexity increases regardless of the classifier used. Note that the complex texture feature, LBP, performs no better than the simple edge feature LoG, and delivers far inferior performances as complex shape and edge features, HoG, and mHoG. We hypothesize that the lack of performance improvement is because the shapes and edges, such as those from the limbus and sclera, communicate more information about the gaze location than texture does. Note that, mHoG and HoG achieve the best results and mHoG performs slightly better than HoG, while other features yield far worse results. Moreover, the computation of mHoG feature is fast due to the utilization of integral histograms~\cite{integralHistogram}.

The rows of Table~\ref{resultTable} represent different regressors. We notice that the best two results both come from the RF regressor. In addition, RF provides fast prediction results, thus has been widely adopted in real time systems~\cite{RFfast1}. In our experiments, we actually found the results were reasonably stable when using more than 20 trees; we used 100~trees to further improve accuracy. 

Overall, mHoG and RF achieve the lowest error of 3.17~$\pm$~2.10 cm, as listed in Table~\ref{resultTable}. A few example images with high estimation error are shown in Fig. \ref{badPrediction}. Even considering the computational complexity (e.g. for real-time applications), mHoG and RF are still recommended for their relatively fast computation. This is our chosen algorithm for the experiments in the following sections.

\begin{table}[!t]
	\renewcommand{\arraystretch}{1.5}
	\caption{Mean error (cm) for each feature and regressor combination. Note that the combination of mHoG feature and RF regressor achieved the lowest error.}
	\label{resultTable}
	\centering
	\resizebox{\columnwidth}{!}{
		\begin{tabular}{|c|c|c|c|c|c|}
			\hline
			\backslashbox{Regressors \kern-1em}{\kern-1em Features} & Raw pixels & LoG & LBP & HoG & mHoG \\
			\hline
			$k$-NN & 9.26 & 6.45 & 6.29 & 3.73 & 3.69 \\
			\hline
			RF & 7.20 & 4.76 & 4.99 & 3.29 & \underline{\textbf{3.17}} \\
			\hline
			GPR \footnotemark[1] & 7.38 & 6.04 & 5.83 & 4.07 & 4.11 \\
			\hline
			SVR \footnotemark[2] & x & x & x & x & 4.07 \\
			\hline
		\end{tabular}
	}
\end{table}
\footnotetext[1]{Due to training time complexity constraint, GPR is evaluated using three-fold cross validation on data of 15 subjects, which is essentially leave-5-subjects-out.}
\footnotetext[2]{SVR is evaluated only on the optimal feature, which is mHoG. The evaluation process is conducted in the same way as GPR.}

\subsection{Person-Dependent and Person-Independent Performance Comparison}
\label{Person Dependent and Independent Performance Comparison}

In prior appearance-based gaze estimation methods \cite{tanLinearManifold,GPRgaze}, the evaluation process of the algorithm use data from the same subject and same session for both training and testing (person and session dependent). Here, we study only the influence of person-dependency, not session-dependency, on algorithm performance. The analysis of session-dependency is not useful because in daily use, a person's appearance can vary widely between sessions. In a person-dependent model, the performance is evaluated using leave-one-session-out cross validation on the data from the same person (each person has 16 sessions). In the person-independent model, a leave-one-subject-out cross validation is employed. In each one of the 41 evaluation rounds (the TabletGaze dataset includes 41 subjects), the regressor is trained on data from 40 subjects and tested on the remaining one subject, and then the final results are obtained by averaging the estimation errors of all the images from the 41 subjects.

\begin{figure}[!t]
	\centering
	\includegraphics[width=\columnwidth]{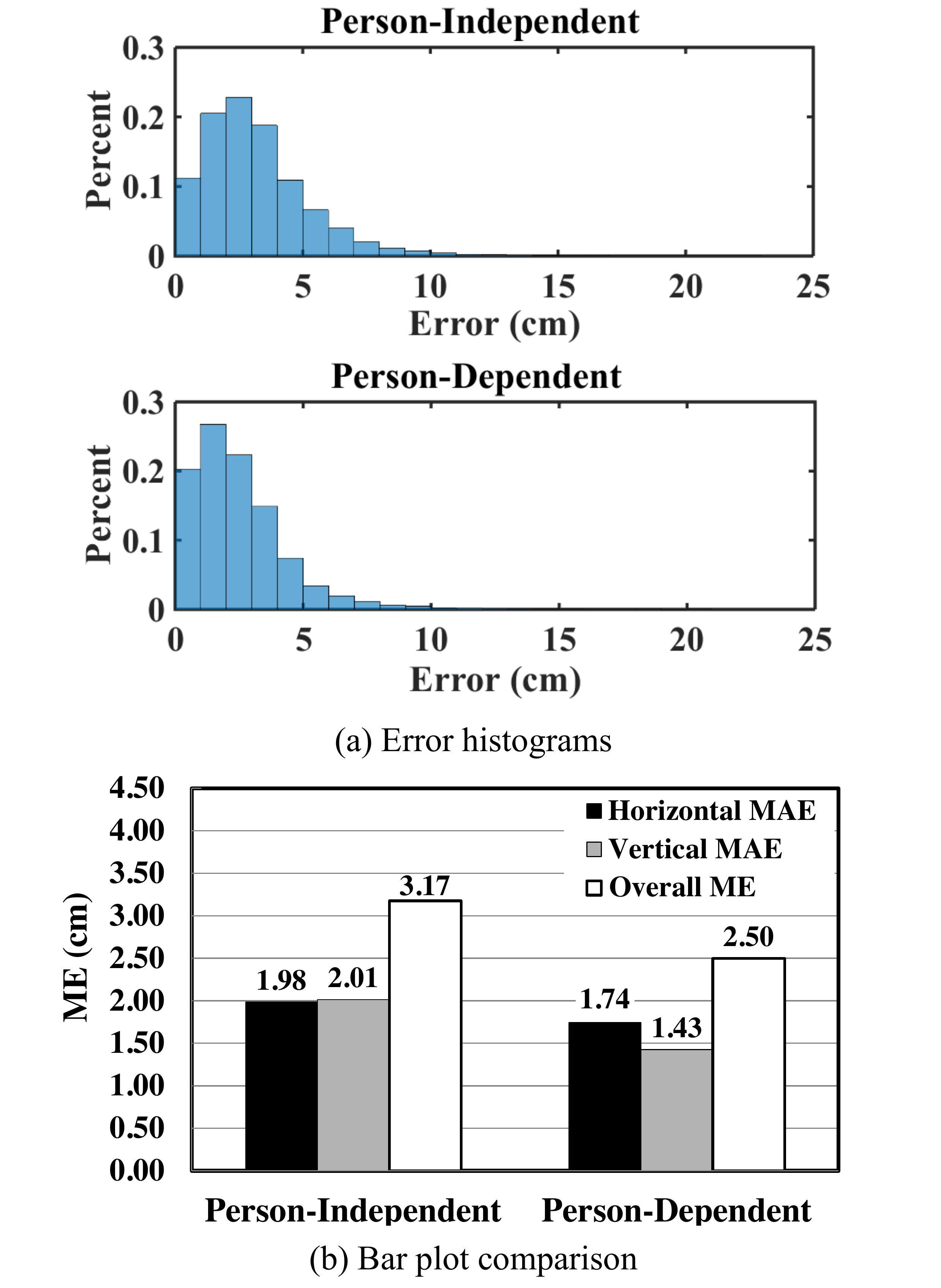}
	\caption{Person-independent and person-dependent training performance comparison. In person-independent training, leave-one-subject-out cross validation was utilized to evaluate the algorithm; while in person-dependent training, leave-one-session-out cross validation was employed.}
	\label{dependentVSindependent}
\end{figure}

Fig. \ref{dependentVSindependent}(a) shows the estimation error histograms over all the images in the sub-dataset. We observe that for person-dependent training scenario, the estimation errors aggregate near lower values compared to person-indpendent training scenario. The observation implies that for the person-dependent training scenario, the estimation error is lower than that in the person-independent training scenario. The numerical MEs over all samples in the sub-dataset are shown in Fig. \ref{dependentVSindependent}(b) for the two training scenarios. This result is expected because the regressor will have better generalization power for images from the same person, due to the stronger similarity between the images.

We also present the stand alone horizontal and vertical errors (x and y coordinates on the tablet screen), in addition to the overall/combined ME for both person-dependent and person-independent training scenarios in Fig. \ref{dependentVSindependent}(b). The horizontal and vertical errors are both evaluated using mean absolute error (MAE) to avoid the cancellation of positive and negative errors. Unidirectional gaze estimation might be useful for applications that requires only information from a singular direction, such as web-page scrolling. We observe that the horizontal and vertical errors are similar, showing that the horizontal and vertical regressors have similar predictive powers. 

\subsection{Comparison with Prior Results}
\label{Comparison with Prior Results}

In this section, we compare our results with those reported in previous works. In order to conduct this comparison, we convert our distance error into angular error. We find that the distance between the subject's eyes and the tablet varied from 30 to 50 cm. For the sake of comparison, we compute the error for that range of distances. Given the error distance on tablet screen $E$ and the tablet-to-eyes distance $D$ , the angular error of the algorithm is derived by calculating the arctangent of the ratio of the error distance to the distance from the user to the tablet $arctan(\frac{E}{D})$. 

The first work against which we compare our results is the work done in~\cite{eyetab} for mobile tablets using a geometry-based gaze estimation approach. Note that the authors used 9 gaze locations covering only part of the tablet screen. Moreover, the data was collected with a fixed user-tablet distance and the result was reported for a person-dependent study. 
The second comparative work \cite{wildGazeEstimation} is a study of appearance-based gaze estimation for laptop screens. While the participants for the study used the laptops freely, the variation of user posture on laptops is lower than that on tablets. In addition to eye appearance, head pose was utilized as an input to the algorithm because the full face is visible in the image frames at all times. Our tablet usage environments draw a sharp contrast to the laptop usage environments, since the face is not entirely visible 69.2\% of the time for a tablet user, and thus eliminating the possibility of reliable head pose estimation.
Finally, work done in \cite{ALRegression} claimed a state-of-the-art estimation accuracy for the condition of sparse training samples. In their experiments, they used fewer than 40 training samples per person for a person-dependent training scenario. Their algorithm was evaluated both on data of fixed frontal head pose and on data of slight head motion for 8 subjects. However their method is not easily scalable to large data problems, and they also focused only on person-dependent training.

\begin{figure}[!t]
	\centering
	\includegraphics[width=\columnwidth]{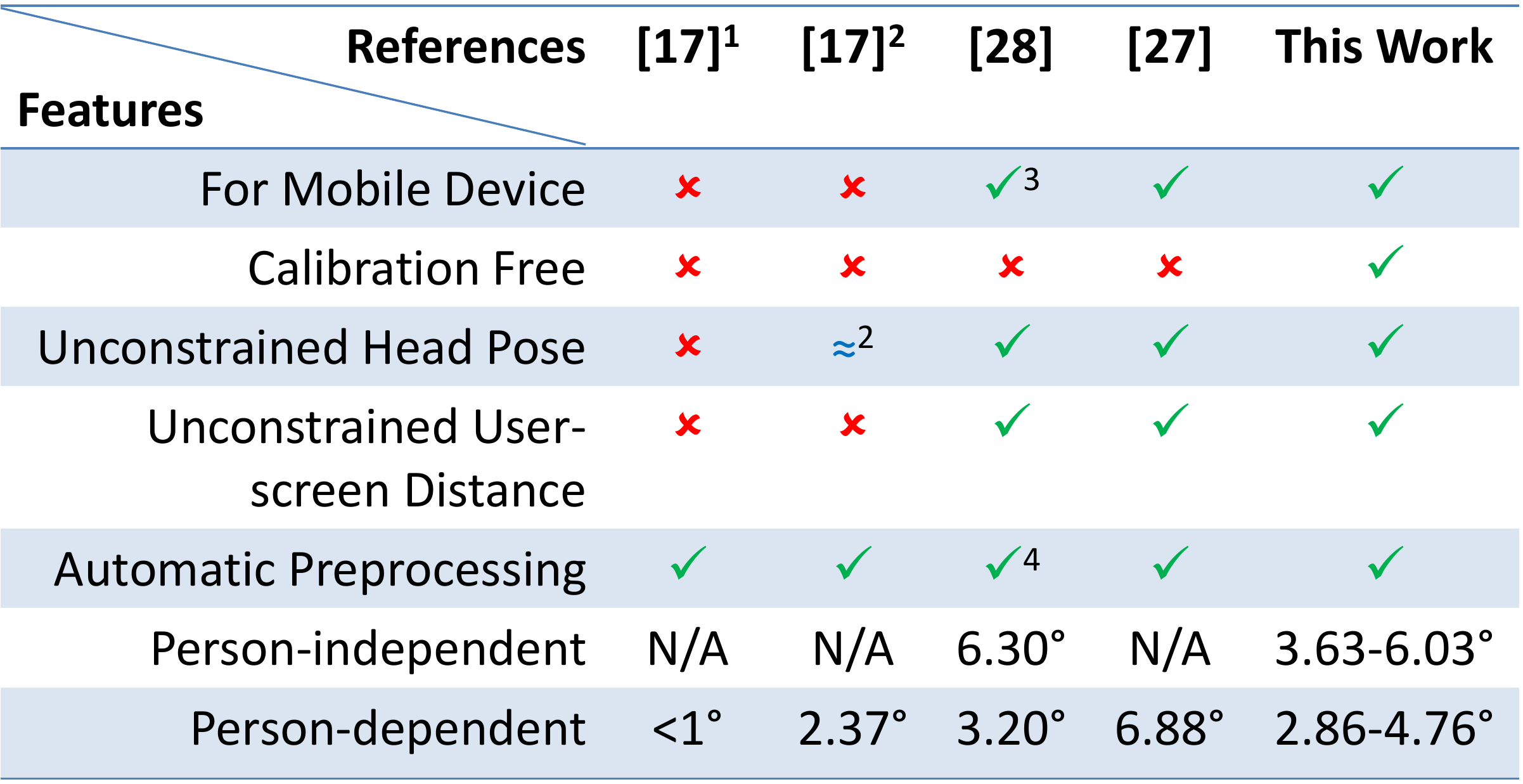}
	\caption{Comparison with prior works. We can observe that our method is free of all the constraints. Moreover, our method achieves better accuracy than Wood et. al \cite{eyetab}, and comparative accuracy than other works both for person-dependent and person-independent trainings.}
	\label{comparisonTable}
\end{figure}

A detailed comparison between our proposed technique and prior works is shown in Fig. \ref{comparisonTable}. In addition to quantitative error, other relevant properties, such as whether requires calibration and restricts head motion, are also listed. We can observe that our method is the only one that is free of all the constraints. Moreover, our method achieves an angular error competitive with other works for both person-dependent and person-independent training scenarios.

\footnotetext[1]{The head pose is fixed (no head motion).}
\footnotetext[2]{The head motion is limited to several degrees.}
\footnotetext[3]{The data is collected for a laptop.}
\footnotetext[4]{The method requires a mean facial shape model built across all subjects.}

\begin{figure}[!t]
	\centering
	\includegraphics[width=\columnwidth]{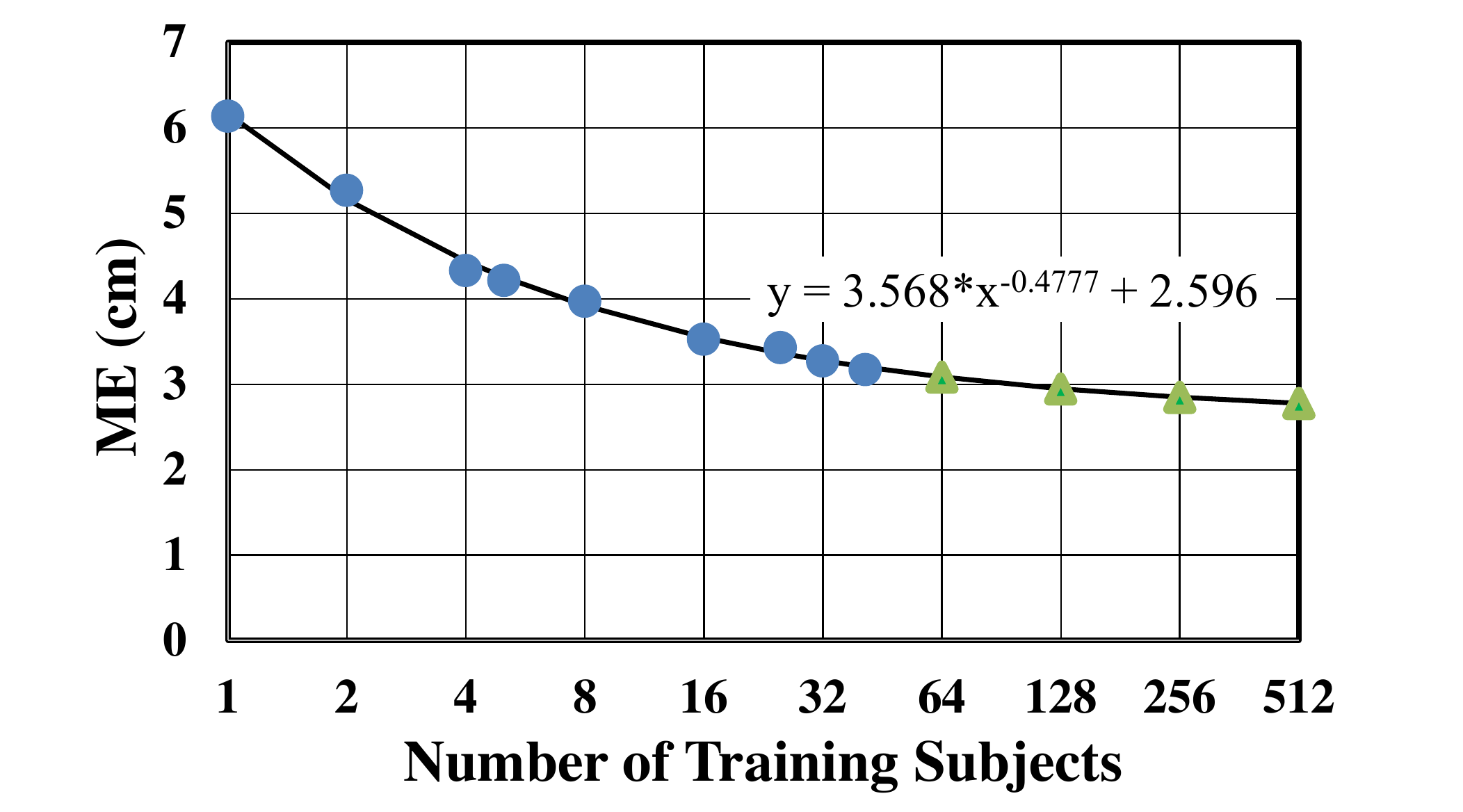}
	\caption{Effect of training data size on the gaze estimation accuracy of TabletGaze. The round circles are results obtained in the experiment, and a line is fitted to the data points. The triangles are data points derived through extrapolation on the fitted line. We can see that the ME decreases monotonically as the number of training subjects grows larger, indicating that more data could improve the performance further.}
	\label{trainingDatasize}
\end{figure}

\subsection{Effect of Training Data Size}
\label{Effect of Training Data Size}

In this section, we study the impact of training data size on the estimation accuracy of TabletGaze. We randomly select groups of different number of participants for evaluation. We experiment with groups of different sizes $K$, where $K$ is within the range $[2, 41]$. For each group, we perform leave-one-subject-out cross validation, so in each training round we use $K-1$ subjects' data. Since we are randomly selecting a subset of data from the whole data, we repeat the same process 5~times and average the final reported errors to reduce bias. 

The results are presented in a semi-log plot as shown in Fig. \ref{trainingDatasize}. As the size of the training group increases, the estimation error decreases monotonically. The monotonically decreasing relationship suggests that if we use more training subjects, we can further improve estimation accuracy.

\subsection{Eyeglasses, Race and Posture}
\label{Eyeglasses, Race and Posture}

We validate whether dividing the dataset into groups based on person-related factors and training a separate regressor for each group would further reduce the estimation error. Our hypothesis is that the eye appearance variations caused by factors other than gaze can be reduced within each group. Previous works on head pose estimation \cite{poseDetectorArray, poseDetectorArray2} and face detection \cite{faceDetectorPyramid} demonstrated improved accuracies by dividing the data into groups and training a regressor/detector for each group. At the same time, we also examine the impact of each factor on gaze estimation accuracy. Due to a lack of sufficient data in some of the categories, for example we have only six subjects who are Caucasians and wearing glasses, we could not perform controlled tests to study the impact of each independent factor. Nevertheless, we can still gain some initial understanding of the impact of the three factors on the performance of the gaze estimation algorithm. Three factors are considered in our study: eyeglasses (wearing eyeglasses or not), race (Caucasian or Asian), and body posture (standing, sitting, slouching or lying). Three experiments are conducted for each factor. 

\subsubsection{Eyeglasses} \label{Eyeglasses}

\begin{figure}[!t]
	\centering
	\includegraphics[width=\columnwidth]{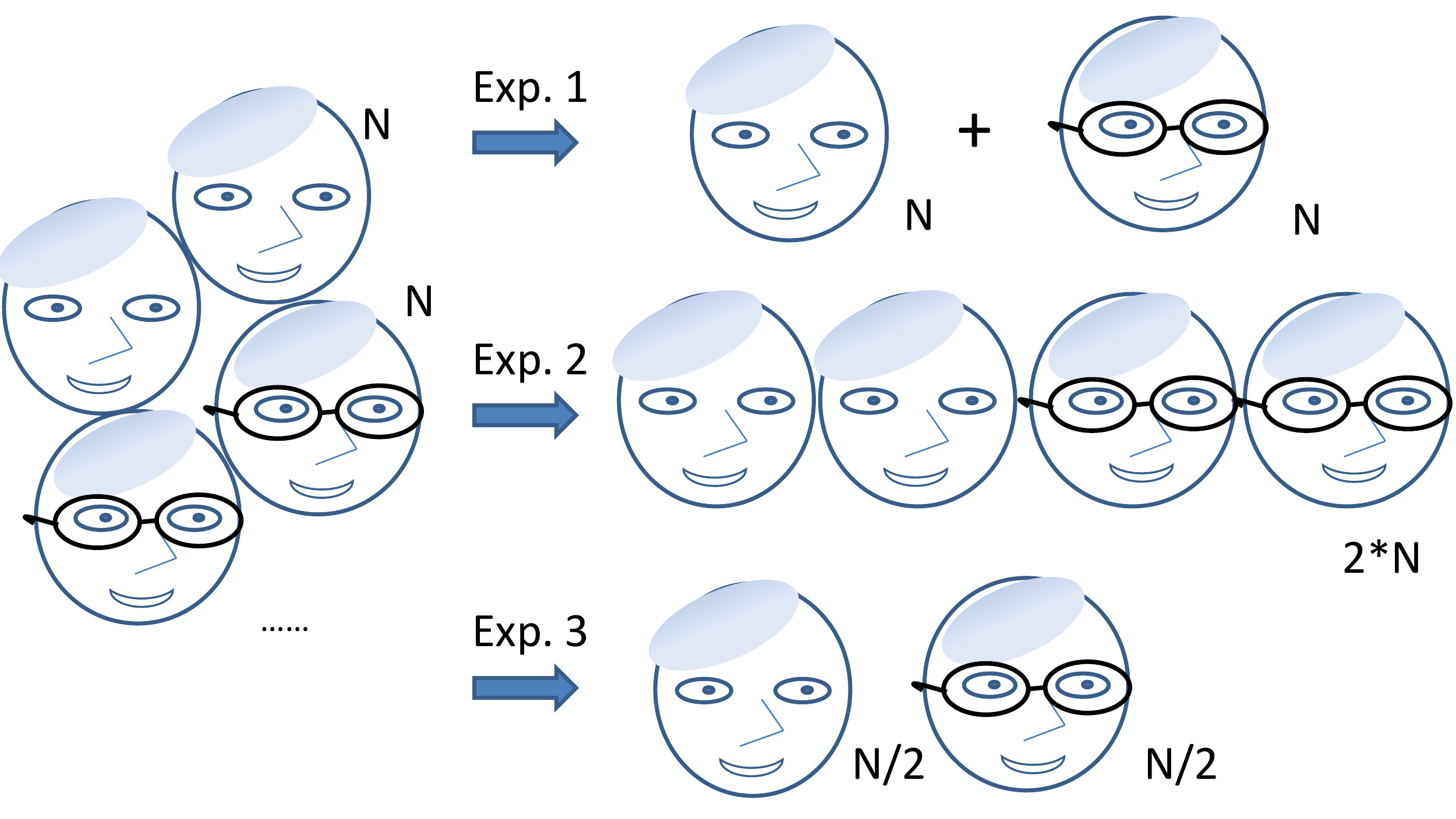}
	\caption{Diagram of the design for the three experiments studying the factor of prescription glasses. In Experiment 1, the dataset was partitioned into 2 groups of wearing glasses (Group 1, $N$ subjects) and not wearing glasses (Group 2, $N$ subjects), and training and testing were done separately for each group. In Experiment 2, the leave-one-subject-out cross validation were conducted on all data ($2 \times N$ subjects), but the ME was separated for each group. In Experiment 3, we combined data of half of the subjects from Group 1 and half from Group 2, and conducted training and testing within the combined data. }
	\label{experimentDesign}
\end{figure}

\begin{figure*}[!t]
	\centering
	\includegraphics[width=\textwidth]{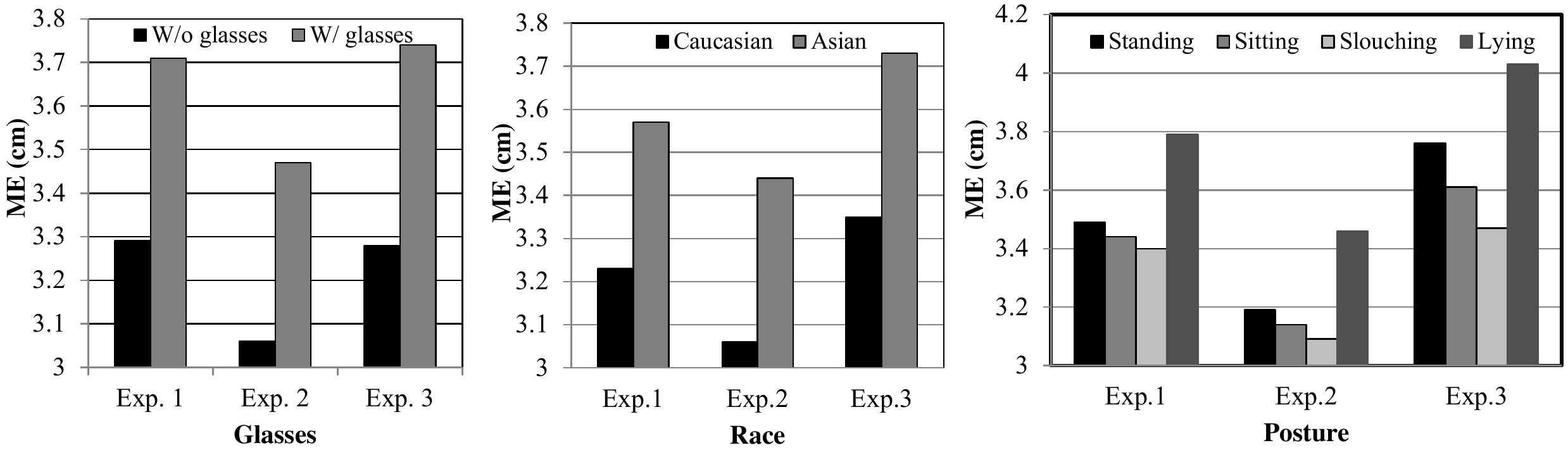}
	\caption{Study of whether partitioning the data based on person-related factors would reduce estimation error. The error obtained in Exp. 2 is lower than that in Exp.1 for all the three factors. It means that when we have limited training subjects, data partition increases the estimation accuracy. The error obtained in Exp.3 is higher than that in Exp. 1 for racial characteristics and body posture, and almost the same for wearing glasses or not. It infers that when we have sufficient training subjects, data partition based on the factor of race and boy posture improves the estimation accuracy, while the factor of glasses does not significantly impact the result.}
	\label{factors}
\end{figure*}

We first discuss the impact of eyeglasses. A diagram of the experiment design for the three experiments is shown in Fig. \ref{experimentDesign}. The dataset is first divided into two groups: Group~1 is wearing glasses, and Group~2 is not. In the first experiment, leave-one-subject-out cross validation is evaluated on the data of each group separately, and the estimation errors are obtained for each group. In our data, there is an unequal number of subjects within each group. To solve this problem, suppose Group~1 has $M$ subjects and Group~2 has $N$ subjects, where $M$ is larger than $N$. Then we randomly select $N$ subjects from Group~1 and run Experiment~1. We repeat the experiment 5 times and average the ME for Group~1 to reduce bias caused by random selection. The second experiment is conducted on data from both groups using leave-one-subject-out cross validation. The estimation error is separated depending on whether the test subjects are wearing glasses or not. In Experiment~1, the number of training subjects is smaller than the number of training subjects in Experiment~2 due to data partitioning. We can infer that this size discrepancy will have a negative impact on the estimation accuracy, as discussed in Section \ref{Effect of Training Data Size}. To mitigate the effects of training data size, in Experiment~3 we choose the same training data size as in Experiment~1. We randomly select $N/2$ subjects from Group~1 and $N/2$ subjects from Group~2, and combine the data in Experiment~3. The evaluation process is done using the same method as in Experiment~2. Experiment~3 is also repeated 5 times to reduce the bias caused by the random selection of training subjects. 

The results are shown in the first bar plot of Fig.~\ref{factors}. As we can observe from the bar plot, the ME of the group of wearing glasses is larger than the group of not wearing glasses for all the three experiments. We can also observe that in Experiment 1, the ME increases around 0.4 cm for the group of wearing glasses compared to the group of not wearing glasses. These observations means that wearing glasses has a negative impact on gaze estimation accuracy. We can tell from the bar plot that for each group, the ME of Experiment~1 is higher than the ME of Experiment~2. The increase of error means that partitioning the data does not improve accuracy when we have limited number of training subjects. We obtain similar ME for Experiments~1 and~3, showing that partitioning the data based on the factor of glasses does not have a significant impact on estimation accuracy when we have sufficient training data. The factor of glasses does not affect estimation accuracy most likely because sometimes the reflection from glasses is not strong and does not introduce much noise in the eye images.

\subsubsection{Race} \label{Race}
We utilize the same approach as in Section \ref{Eyeglasses} to design the three experiments to study the impact of racial characteristics. The second bar plot of Fig.~\ref{factors} shows the results. We obtain quite different MEs for the group of Caucasians and the group of Asians, which tells that the factor of race impacts the performance of the gaze estimation algorithm. We also notice that for each individual group, the ME of Experiment~1 is higher than the ME of Experiment~2 while the ME of Experiment~1 is lower than that of Experiment~3. We can infer that that partitioning the data does not improve accuracy when we have limited number of training subjects. Moreover, when we have a large amount of training data, dividing the data based on raceimproves accuracy because people within the same racial group have similar eye shapes.

\subsubsection{Body Posture} \label{BodyPosture}
For studying the impact of body posture (standing, sitting, slouching or lying), the data partition is performed differently. As described in Section \ref{Data Collection}, we have four subsets of data for each posture for each subject. We partition the dataset into 4~groups, each group containing data of one body postures from the same subjects. For each subject, the data size for each body posture may be unequal due to occasional unusable data. To reduce the effect of unequal dataset size, we choose a subset of 29~subjects from the TabletGaze dataset where the amount of each subject's data for each body posture is almost equal. Then we perform the three experiments in the same way as described earlier in this section.

The results are shown in the third bar plot of Fig.~\ref{factors}. We notice that the MEs of the standing, sitting and slouching groups are quite similar, while the ME for the group of lying is the highest. One reason for the high error of the group of lying is that people have more varied head pose and way of holding the tablet when they are lying. We also notice that for each individual group, the ME of Experiment~1 is higher than the ME of Experiment~2 while the ME of Experiment~1 is lower than that of Experiment~3. We can infer that that partitioning the data does not improve accuracy when we have limited number of training subjects. Moreover, when we have a large amount of training data, dividing the data based on body posture improves accuracy because people might have similar head poses when they are doing the same posture.

\subsection{Continuous Gaze Tracking from Videos}

\begin{figure}[!t]
	\centering
	\includegraphics[width=\columnwidth]{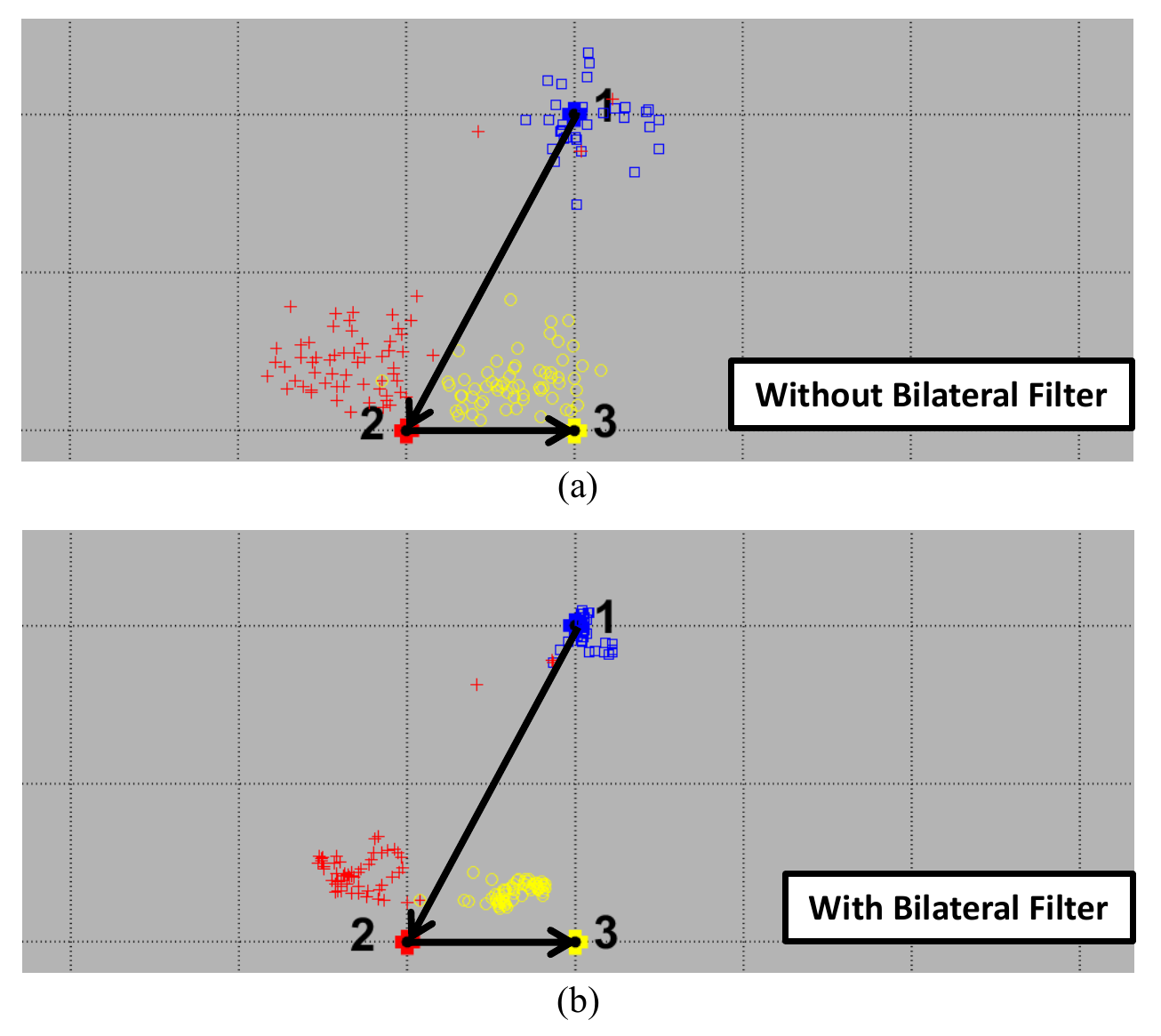}
	\caption{Continuous gaze tracking demonstration. Each image shows a part of the tablet screen, and the cross-stitches of the grid lines represent the 35 potential ground truth gaze locations. In (a), 3 ground truth gaze locations and the color-coded predicted locations are shown for a single subject from our dataset. Each location is showed in the sequence indicated by the arrows. The distance between the predicted gaze locations and the true gaze location is within the distance between two cross-stiches (3.4 cm). In (b), the predicted locations are passed through a bilateral filter; the fluctuations of the predictions are reduced by the filter.}
	\label{continuousGazeTracking}
\end{figure}

We apply the TabletGaze algorithm to continuously track user's gaze on videos in the sub-dataset. Initially we directly estimate user's gaze in the videos on a frame-by-frame basis. When implementing a continuous gaze tracking system, temporal information can be utilized to further reduce gaze prediction errors. A temporal bilateral filter can be applied on consecutive gaze estimations to reduce the miniature fluctuation of neighboring gaze estimations caused by model noise, and preserve the large gaze shifts due to change of fixation location. Some example images of the continuous gaze tracking based on our TabletGaze algorithm, and the effect of bilateral filter are displayed in Fig. \ref{continuousGazeTracking}. We can observe that for each gaze location in the image, the gaze estimations are close to the ground truth gaze locations, and the errors are less than 3.4 cm (the distance between two cross-stiches), which conforms to the ME of 3.17cm. We also notice that after applying the bilateral filter, fluctuations of the gaze estimations for each ground truth gaze location are decreased. Meanwhile, temporal eye center location information can be collected, and the change of subsequent eye center locations can be used to correct gaze estimations. For example, sometimes a user naturally moves his/her head from left to right when he/she is looking from left to right on the screen. Along with the head movement, the eye center location would also shift to the right. This shift of eye location can thus be utilized to correct neighboring gaze estimates so the predicted gaze location also changes accordingly.

\section{Discussion and Conclusion} \label{Discussion and Conclusion}
All of the evaluations of the algorithm are conducted on a desktop computer. When implementing the algorithm on a tablet, the RF regressor can be pre-trained off-line and loaded onto the device. The computation of the mHoG feature from an image and prediction using the RF model is fast, which means real-time estimation is possible. An explicit 3D head pose is not utilized in this work. Here we discuss an exploratory experiment regarding incorporating implicit head pose information. 

As we discussed earlier, direct 3D head pose information cannot always be obtained for the mobile environment due to partial facial visibility in some cases. But head pose information is correlated with features such as the location of the eye center in the image frame and the size of the eyes, which can be extracted as alternatives to exact head pose angles. To utilize this information, we design a feature vector composed of the following features: the x and y coordinates of the left and right eyes, eye sizes (width and height of the eye bounding boxes), and the x and y location difference between the left and right eyes. This feature vector has a length of 10 and is combined with the LDA reduced mHoG feature as an input to the RF regressor. The data is also evaluated using leave-one-subject out cross validation, and we obtain a ME of 3.10 $\pm$ 2.07 cm. There is no significant improvement compared to the 3.17 $\pm$ 2.10 cm ME when we do not use the eye location information. This means directly adding these features does not result in significantly improved estimating accuracy. A future direction could be focused on designing a new scheme to appropriately and productively incorporate eye location information.

In conclusion, this work presented and studied the unconstrained mobile gaze estimation problem in three major steps. Firstly, a large dataset was collected in an unconstrained environment. To the best of our knowledge, this is the first dataset of its kind. The dataset is designed to explore the variation of subject appearances in an unconstrained environment by including 4 different postures and recording the data in videos. 3 observations were made on the dataset, including facial visibility, posture, and glasses reflection, which provide a deeper understanding of the challenges present in the mobile environment. An automatic gaze estimation algorithm is presented, composed of currently available computer vision tools, which can estimate a person’s gaze from an image recorded by the tablet front camera. The algorithm achieves a ME of 3.17 $\pm$ 2.10 cm on the tablet screen, which is a significant improvement over prior works on mobile gaze estimation. The result is good for applications that do not require high accuracy on the tablet.

\begin{acknowledgements}
We acknowledge the support from National Science Foundation (NSF) Grants NSF-IIS: 1116718, NSF-CCF:1117939 and NSF-CNS:1429047. We would further like to thank all the participants in the dataset for volunteering and allowing their data to be released.
\end{acknowledgements}

\bibliographystyle{spmpsci}      
\bibliography{manuscript}

\end{document}